%% file: main_arxiv.tex
\documentclass{article} 
\usepackage{iclr2027_conference,times}

\input{math_commands.tex}

\usepackage{hyperref}
\usepackage{url}
\usepackage{graphicx}
\usepackage{booktabs}
\usepackage{multirow}
\usepackage{caption}
\usepackage{placeins}
\usepackage[table]{xcolor}
\definecolor{rocobg}{RGB}{242,247,252}
\usepackage{wrapfig}
\usepackage{needspace}
\usepackage{float}

\newcommand{\olmoe}{OLMoE}
\newcommand{\qwenmoe}{Qwen1.5}
\newcommand{\deepseek}{DeepSeek}
\newcommand{\qwen}{Qwen2.5}
\newcommand{\llama}{Llama}
\newcommand{\qwenthree}{Qwen3}

\title{A Strong MoE Is All You Need\\for Process Supervision}

\author{
\begin{tabular}{@{}l@{}}
{\normalfont\bfseries
Guanzhi Deng\textsuperscript{\rm 1},
Zengyan Liu\textsuperscript{\rm 1},
Haibo Wang\textsuperscript{\rm 2},
Kuan Wu\textsuperscript{\rm 1},
Shing Yin Wong\textsuperscript{\rm 1},
}\\
{\normalfont\bfseries
Sichun Luo\textsuperscript{\rm 3},
Zhuoran Wang\textsuperscript{\rm 1},
Linqi Song\textsuperscript{\rm 1}\thanks{Corresponding author.}
}\\[6pt]
{\normalfont
\textsuperscript{\rm 1}City University of Hong Kong
\quad
\textsuperscript{\rm 2}Carnegie Mellon University
\quad
\textsuperscript{\rm 3}University of Hong Kong
}\\
{\normalfont
\texttt{guanzdeng2-c@my.cityu.edu.hk}
\quad
\texttt{linqi.song@cityu.edu.hk}
}
\end{tabular}
}

\iclrfinalcopy

\makeatletter

\patchcmd{\@maketitle}
  {\lhead{Published as a conference paper at ICLR 2027}}
  {\lhead{Preprint. Under review.}}
  {}{}

\patchcmd{\@maketitle}
  {\@title\par}
  {\@title\par\vskip 0.15in}
  {}{}

\makeatother

\begin{document}

\maketitle

\begin{abstract}
Process reward models (PRMs) provide fine-grained supervision for multi-step reasoning, yet obtaining reliable process labels remains a central bottleneck. Existing approaches typically rely on human annotation, rollout-based estimation, or auxiliary verification. In this work, we explore a different source of process supervision natively exposed by Mixture-of-Experts (MoE) models: expert routing. We show that simple routing statistics provide an intrinsic process-confidence signal that is competitive with conventional output-logit confidence. Building on this observation, we construct routing confidence by combining overall routing mass, head routing mass, and within-step routing variation, and introduce \textbf{RoCo-Rethink}, which identifies the least confident reasoning step and selectively regenerates the subsequent trajectory. We further introduce \textbf{RoCo-PRM}, which uses a strong MoE to automatically annotate reasoning trajectories with routing-derived step scores and distills them into a standalone PRM. Across multiple MoE architectures and diverse reasoning tasks, both RoCo-Rethink and RoCo-PRM consistently outperform strong baselines, while RoCo-PRM transfers effectively across both model architectures and tasks. Our findings suggest that expert routing is more than a computational dispatch mechanism: it provides an architecture-native source of process supervision for reasoning.
\end{abstract}

\section{Introduction}

Large language models (LLMs) have made rapid progress on multi-step reasoning tasks, yet their reasoning trajectories remain highly fallible. A model may proceed correctly for several steps before making an error that propagates through the rest of the solution. This has motivated increasing interest in \emph{process supervision}, where intermediate reasoning states are evaluated rather than assigning feedback only to the final outcome. Process reward models (PRMs) have consequently become an important component of reasoning verification, test-time search, self-correction, and reinforcement learning \citep{lightman2024verify,wang2024mathshepherd,setlur2025rewarding,zheng2026comprehensive}.

A fundamental challenge is where such process supervision should come from. Existing approaches obtain process supervision from human annotations, rollout-based estimation, automated verification, outcome-derived supervision, or model-internal uncertainty signals \citep{lightman2024verify,wang2024mathshepherd,sun2025epicprm,tan2026aurora,zhao2026genprm,yuan2025freeprocess,sun2025freeprm,cao2026eduprm,lee2026cpmi}. While recent automated methods substantially reduce dependence on manual annotation, they often require additional evaluation computation, outcome signals, or auxiliary evaluators. Comparatively little attention has been paid to architecture-native computation signals that arise naturally during a single forward pass. This is particularly relevant for Mixture-of-Experts (MoE) models, where routers continuously map hidden representations to sparse expert allocations. These routing decisions are produced intrinsically during ordinary inference, raising the central question of this work:

\begin{quote}
\emph{Can native MoE routing serve as intrinsic process supervision for reasoning?}
\end{quote}

There are both architectural and empirical reasons to believe that it can. Given a hidden state $h_t$, token prediction and expert routing can be viewed as two distinct projections of the same latent representation:
\begin{equation}
h_t \xrightarrow{W_{\mathrm{LM}}} p(v \mid h_t),
\qquad
h_t \xrightarrow{W_{\mathrm{router}}} p(e \mid h_t).
\end{equation}
The former reflects uncertainty over what token to generate next, whereas the latter reflects how computation is allocated across experts. Recent work further shows that native MoE routing preserves structured information about latent reasoning states beyond the emitted trace \citep{chen2026sametoken,chen2026beyondtrace}. This motivates us to ask whether routing behavior can be compressed into a lightweight process-confidence signal and further converted into reusable process supervision.

We investigate this through \emph{routing confidence}, which combines three complementary routing statistics at each reasoning step: overall routing mass, head routing mass, and within-step routing variation. Across multiple MoE architectures and reasoning domains, these statistics exhibit systematic relationships with step-level reliability, and routing-based predictors often outperform conventional output-side confidence signals.

Building on this signal, we introduce \textbf{RoCo} (\textbf{Ro}uting \textbf{Co}nfidence), with two instantiations. \textbf{RoCo-Rethink} directly uses routing confidence at test time to identify the least confident reasoning step, preserve the preceding prefix, and regenerate the subsequent trajectory. \textbf{RoCo-PRM} uses a strong MoE teacher to annotate reasoning trajectories with routing-derived step scores and distills them into a standalone PRM that scores reasoning steps without access to routing information.

Empirically, RoCo-Rethink consistently outperforms strong test-time reasoning baselines, while RoCo-PRM consistently surpasses automatic PRM baselines, including when applied to dense target models. Further analyses show that composite routing confidence more reliably identifies revision points with high corrective value and that RoCo-PRM benefits more from stronger routing teachers than from larger PRM backbones.

Our main contributions are fourfold.
(1) We introduce a lightweight step-level routing-confidence formulation that combines three complementary statistics from native MoE routing.
(2) We develop RoCo-Rethink, which uses routing confidence to identify error-prone reasoning steps and selectively regenerate the subsequent trajectory.
(3) We introduce RoCo-PRM, which uses a strong MoE to automatically annotate reasoning trajectories with routing-derived step scores and distills them into a standalone PRM that can guide both MoE and dense target models.
(4) RoCo-Rethink outperforms the strongest test-time baseline in all 18 evaluated settings, while RoCo-PRM achieves the best or tied-best performance in all 30 settings.


\section{Related Work}

\subsection{Process Supervision and Process Reward Models}

Process supervision provides fine-grained feedback on intermediate reasoning steps and has become an important paradigm for improving multi-step reasoning. Early work demonstrates the effectiveness of human-annotated process supervision \citep{lightman2024verify}, while subsequent studies construct process supervision automatically through rollouts, outcome-derived signals, automated verification, or model-internal uncertainty \citep{wang2024mathshepherd,setlur2025rewarding,sun2025epicprm,yuan2025freeprocess,sun2025freeprm,tan2026aurora,zhao2026genprm,cao2026eduprm,lee2026cpmi}. Many of these approaches nevertheless require additional evaluation computation, outcome information, or auxiliary evaluators. In contrast, we investigate whether process supervision can be derived directly from architecture-native routing dynamics arising during ordinary MoE inference.

\subsection{MoE Routing as a Reasoning-State Signal}

Mixture-of-Experts models dynamically route tokens to sparse subsets of experts, with prior work largely focusing on routing efficiency, load balancing, and expert specialization \citep{fedus2022switch,dai2024deepseekmoe,muennighoff2025olmoe}. More recent studies suggest that routing also reflects latent reasoning dynamics. \citet{chen2026sametoken} show that routing retains information about task context and trajectory state even when emitted tokens are controlled, while \citet{chen2026beyondtrace} demonstrate that native routing can recover structured representations of latent reasoning states. Complementary studies further connect routing behavior with reasoning reliability and the consequences of routing interventions \citep{yoon2026misrouted,xu2026seeing}. Building on this emerging view, we ask whether routing information can be compressed into a lightweight step-level confidence signal and further converted into reusable process supervision.

\subsection{Test-Time Reasoning Guidance}

Test-time reasoning methods improve model performance by allocating additional computation to uncertain or difficult parts of a reasoning trajectory. General self-correction frameworks iteratively critique, revise, or regenerate model outputs \citep{madaan2023selfrefine,shinn2023reflexion,gou2024critic}, while more recent work uses localized confidence or process verification to determine when and where additional reasoning is needed \citep{yao2024learning,chang2025step,lee2026steer,jin2026corefine}. These approaches demonstrate the value of fine-grained reliability signals for adaptive reasoning guidance, but typically rely on output-side confidence or separately trained evaluators. In contrast, RoCo-Rethink derives its guidance signal directly from native MoE routing and uses it to identify low-confidence reasoning steps for targeted regeneration.

\section{Understanding MoE Routing as a Process Signal}
\label{sec:routing_signal}

We first examine whether native MoE routing tracks step-level reasoning reliability and provides a competitive alternative to conventional output-side confidence. Based on these observations, we construct a lightweight routing-confidence signal for reasoning guidance and process supervision.

\subsection{Routing Signals Track Step-Level Reliability}
\label{sec:routing_reliability}

Inspired by the step-confidence decomposition of \citet{yao2024learning}, we examine three complementary dimensions of routing behavior: overall routing mass ($M$), head routing mass ($H$), and within-step routing variation ($D$).

For token $t$ at MoE layer $\ell$, let
\begin{equation}
    r_{t,\ell}
    =
    \sum_{e \in \mathrm{TopK}_{t,\ell}}
    p_{t,\ell}(e)
\end{equation}
denote the routing probability mass assigned to the selected top-$k$ experts. To account for scale differences across layers, we min--max normalize $r_{t,\ell}$ within each MoE layer over the current reasoning trajectory and then average across the $L$ MoE layers, yielding a token-level routing signal $r_t$.

For reasoning step $s_i$ with token set $\mathcal{T}_i$ and head-token subset $\mathcal{H}_i \subseteq \mathcal{T}_i$, we define
\begin{equation}
    M_i
    =
    \frac{1}{|\mathcal{T}_i|}
    \sum_{t \in \mathcal{T}_i} r_t,
    \qquad
    H_i
    =
    \frac{1}{|\mathcal{H}_i|}
    \sum_{t \in \mathcal{H}_i} r_t,
    \qquad
    D_i
    =
    \operatorname{Std}_{t \in \mathcal{T}_i}(r_t).
\end{equation}
Thus, $M_i$ captures the overall routing mass of the step, $H_i$ emphasizes routing behavior near the beginning of the step, and $D_i$ measures within-step routing variation.

\begin{figure}[t]
    \centering
    \includegraphics[width=\linewidth]{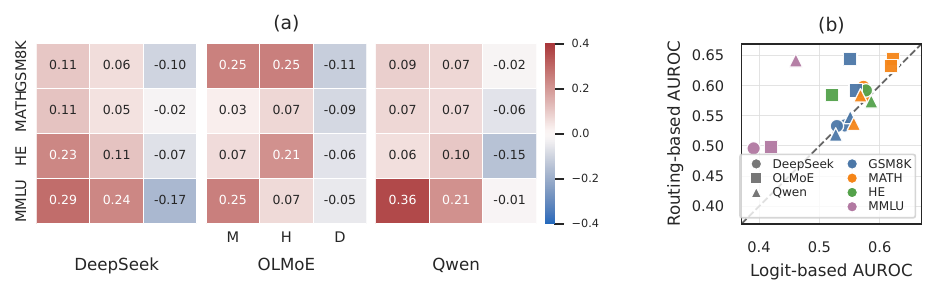}
    \caption{
    \textbf{Native MoE routing tracks step-level reliability and is competitive with output-side confidence.}
    (a) Within-trajectory Spearman correlations between verifier-assessed step quality and overall routing mass ($M$), head routing mass ($H$), and within-step routing variation ($D$) across three MoE models and four tasks. GSM8K and MATH results are averaged over two math PRMs.
    (b) Grouped-CV bad-step AUROC of routing-based versus logit-based predictors. Each point denotes one model--task--verifier setting; points above the diagonal indicate stronger predictive utility from routing features.
    }
    \label{fig:routing_signal}
\end{figure}

We first ask whether these routing statistics systematically track intermediate reasoning quality. Using within-trajectory Spearman correlation, we compare their relative ordering with scores from external process verifiers across three MoE architectures and multiple reasoning domains. For mathematical tasks, we use two independently trained math PRMs as references \citep{zhang2025lessons,xiong2024implementation}; for MMLU and code reasoning, we use VersaPRM and CodePRM, respectively \citep{zeng2025versaprm,li2025codeprm}. As shown in Figure~\ref{fig:routing_signal}(a), higher $M$ and $H$ are associated with higher step quality in nearly all settings, whereas greater $D$ exhibits the opposite relationship. The same directional pattern persists across independent verifiers and reasoning domains, indicating that native routing statistics preserve information about step-level reliability.

We then examine whether this routing information is competitive with conventional output-side confidence. We train lightweight bad-step predictors using either routing features or logit features and evaluate them with grouped cross-validation at the reasoning-trajectory level. Figure~\ref{fig:routing_signal}(b) shows that routing-based predictors outperform their logit-based counterparts in 13 of 18 model--task--verifier settings. Together, the two analyses show that native routing not only tracks intermediate reasoning quality, but also provides a reliability signal that is often stronger than conventional output-side confidence.

\subsection{From Routing Statistics to Routing Confidence}
\label{sec:routing_confidence}

Given that $M$ and $H$ consistently correlate positively with step reliability while $D$ shows the opposite trend, we combine them into a lightweight step-level confidence score. We min--max normalize each statistic across reasoning steps within the current trajectory, yielding $\widetilde{M}_i$, $\widetilde{H}_i$, and $\widetilde{D}_i$ for step $s_i$, and define
\begin{equation}
    C_i
    =
    \widetilde{M}_i
    +
    \widetilde{H}_i
    -
    \widetilde{D}_i.
\end{equation}

The signs follow the empirical pattern in Figure~\ref{fig:routing_signal}(a): greater overall and head routing mass indicate higher reliability, whereas greater within-step routing variation indicates lower reliability. Importantly, routing confidence is computed entirely from quantities already produced during ordinary MoE inference, requiring neither additional rollouts nor an auxiliary verifier.

\section{Method}
\label{method}

\begin{figure*}[t]
    \centering
    \includegraphics[width=\textwidth,
     trim=0cm 6cm 0cm 4.5cm,
    clip]{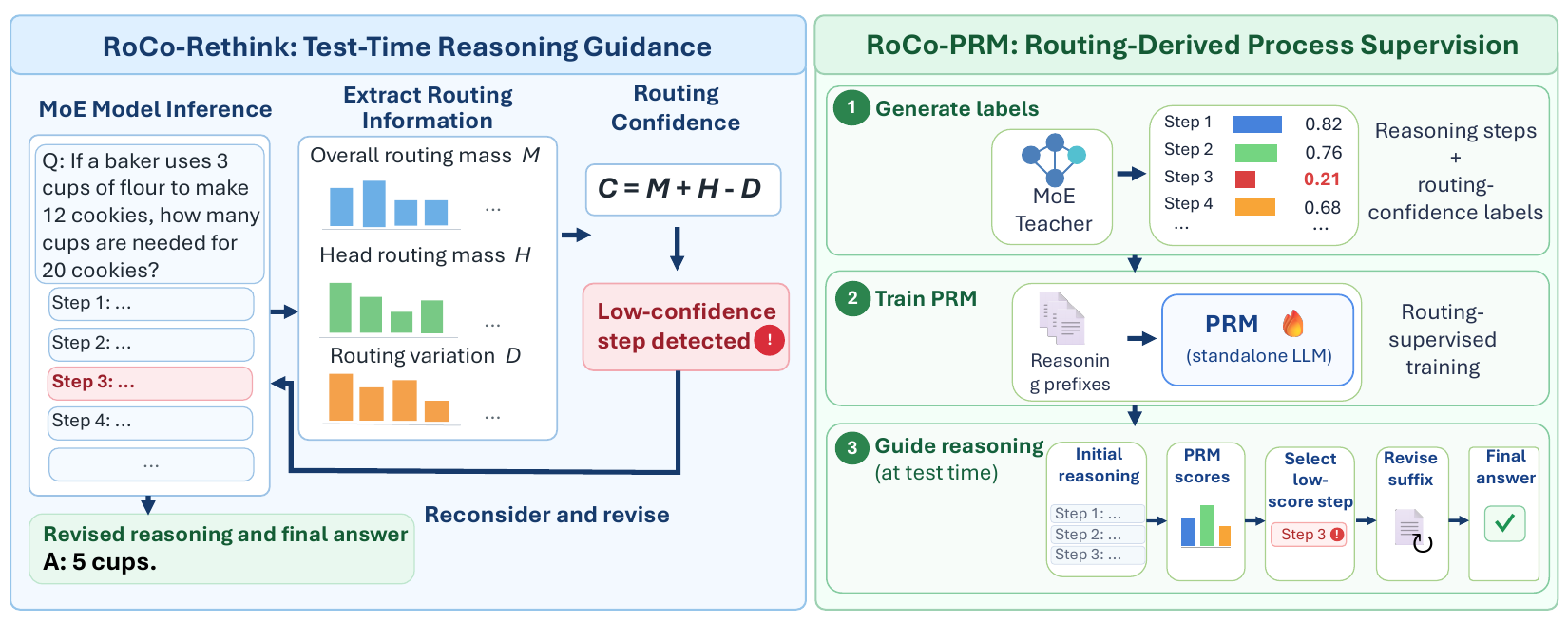}
    \caption{
    \textbf{Overview of RoCo.}
    RoCo leverages native MoE routing confidence for both test-time reasoning guidance and automatic process supervision.
    RoCo-Rethink uses routing confidence to identify and revise low-confidence reasoning steps, while RoCo-PRM distills routing-derived supervision into a standalone PRM that guides reasoning across target models.
    }
    \label{fig:roco_overview}
\end{figure*}

Building on the routing-confidence signal defined in Section~\ref{sec:routing_confidence}, we introduce \textbf{RoCo} (\textbf{Ro}uting \textbf{Co}nfidence; Figure~\ref{fig:roco_overview}), a framework that turns native MoE routing into step-level reasoning guidance and process supervision. RoCo has two instantiations: RoCo-Rethink directly uses routing confidence for test-time reasoning guidance, while RoCo-PRM distills routing-derived supervision into a standalone process evaluator.

\FloatBarrier

\subsection{RoCo-Rethink}
\label{sec:routing_rethink}

Given a reasoning trajectory
\begin{equation}
    S=(s_1,s_2,\ldots,s_T)
\end{equation}
with routing-confidence scores $\{C_i\}_{i=1}^{T}$, RoCo-Rethink selects the least confident reasoning step as
\begin{equation}
    i^{*}
    =
    \arg\min_i C_i.
\end{equation}
Rather than regenerating the entire response, we preserve the reasoning prefix before $s_{i^{*}}$ and regenerate the trajectory from that point onward. Specifically, the model receives the original input together with the preserved prefix $(s_1,\ldots,s_{i^{*}-1})$ and generates a new continuation beginning from step $s_{i^{*}}$. In this way, additional test-time computation is selectively allocated to the portion of the trajectory indicated as least reliable by routing confidence.

The procedure is applied iteratively. After each regeneration, we recompute routing confidence over the updated trajectory and select the currently least confident step for the next rethink round. We terminate when the prediction stabilizes across successive rounds or when a predefined maximum number of rethink rounds is reached. The final output is determined using task-appropriate aggregation or convergence criteria, detailed in Appendix~\ref{app:rethink_details}. In this way, RoCo-Rethink turns native routing information into a direct step-selection signal for targeted reasoning revision without requiring a separately trained process evaluator.

\subsection{RoCo-PRM}
\label{sec:routing_prm}

RoCo-PRM uses routing confidence as an automatic source of process supervision. Given a sufficiently capable MoE teacher, we generate reasoning trajectories and assign each intermediate step $s_i$ its routing-confidence score $C_i$. This produces automatically labeled training examples of the form
\begin{equation}
    (x,s_{\leq i},C_i),
\end{equation}
where $x$ denotes the original problem and $s_{\leq i}$ denotes the reasoning prefix up to step $i$.

We then train a standalone PRM to predict the routing-derived confidence score from the problem and reasoning prefix. Let $f_{\theta}(x,s_{\leq i})$ denote the predicted score for step
$i$, and let $i_n$ denote the step index associated with the $n$-th
training example. We optimize the PRM by regression toward the
routing-derived target:
\begin{equation}
    \mathcal{L}_{\mathrm{PRM}}
    =
    \frac{1}{N}
    \sum_{n=1}^{N}
    \left(
        f_{\theta}
        \left(
            x^{(n)},
            s^{(n)}_{\leq i_n}
        \right)
        -
        C_{i_n}^{(n)}
    \right)^2.
\end{equation}
Because routing information is used only by the MoE teacher to construct supervision labels, the resulting PRM need not itself be an MoE and does not require routing access from the target model.

At test time, RoCo-PRM follows the same step-level reasoning-guidance principle as RoCo-Rethink, but replaces direct routing confidence with the learned PRM score. Given a target-model trajectory, the PRM scores each reasoning prefix, the lowest-scoring step is selected as the revision point, and the target model regenerates the subsequent trajectory conditioned on the preserved prefix. This allows the same PRM to guide both MoE and dense target models, converting architecture-native routing information into reusable process supervision that is decoupled from routing access.

\section{Experiments}
\label{experiments}

\subsection{Experimental Setup}

\textbf{Models and Benchmarks.}
For RoCo-Rethink, we evaluate three MoE models: DeepSeek-V2-Lite-Chat \citep[][hereafter \deepseek]{liu2024deepseek}, OLMoE-1B-7B-0924-SFT \citep[][\olmoe]{muennighoff2025olmoe}, and Qwen1.5-MoE-A2.7B-Chat \citep[][\qwenmoe]{qwen2024qwen15moe}. Each model derives routing confidence from its native expert-routing behavior for selective reasoning revision. For RoCo-PRM, we evaluate the resulting PRM on these three MoE models and two dense models, Qwen2.5-7B-Instruct \citep[][\qwen]{qwen2025qwen25} and Llama-3.1-8B-Instruct \citep[][\llama]{dubey2024llama3}, allowing us to assess whether routing-derived supervision transfers beyond MoE architectures.

We evaluate six benchmarks covering mathematical reasoning, knowledge reasoning, code generation, instruction following, and logical reasoning: GSM8K~\citep{cobbe2021training}, MATH500~\citep{hendrycks2021measuringmath}, MMLU~\citep{hendrycks2021mmlu}, HumanEval~\citep{chen2021evaluating}, IFEval~\citep{zhou2023instruction}, and LogicBench~\citep{parmar2024logicbench}. For PRM training, supervision data are constructed from training datasets that are disjoint from the corresponding evaluation sets, with dataset-specific sources detailed in Appendix~\ref{app:implementation}.

\textbf{Baselines.}
For test-time reasoning guidance, we compare RoCo-Rethink with Vanilla decoding, Self-Consistency~\citep{wang2023selfconsistency}, and LECO~\citep{yao2024learning}. Self-Consistency aggregates multiple sampled reasoning trajectories, while LECO uses generation-logit-based confidence to identify low-confidence steps for selective regeneration. For process supervision, we compare RoCo-PRM with MC-PRM~\citep{wang2024mathshepherd}, which estimates step quality through Monte Carlo rollouts, and CPMI-PRM~\citep{lee2026cpmi}, which derives process rewards from contrastive mutual information between reasoning steps and target answers.

\textbf{Implementation Details.}
All experiments are repeated with five random seeds, and we report the mean and standard deviation. For RoCo-Rethink, the initial response is generated greedily and subsequent rethink rounds regenerate from the least confident reasoning step. For RoCo-PRM, we use Qwen3-30B-A3B~\citep{yang2025qwen3technicalreport} as the routing teacher and Qwen2.5-7B-Instruct as the PRM backbone. At evaluation time, all learned PRMs use the same step-level rethink protocol: the lowest-scoring step is selected as the revision point, and the target model regenerates the suffix from the preserved prefix. Detailed implementation settings are provided in Appendix~\ref{app:implementation}.

\subsection{Main Results}
\label{sec:main_results}

Table~\ref{tab:main_results} summarizes the main results.

\paragraph{RoCo-Rethink consistently improves test-time reasoning guidance.}
RoCo-Rethink achieves the best performance in all 18 model--task settings across the three MoE models. On average, it improves over the strongest test-time baseline by 1.2, 1.9, and 1.5 points on OLMoE, DeepSeek, and Qwen1.5, respectively. Compared with Vanilla decoding, the corresponding average gains are 4.5, 4.8, and 4.5 points. The improvements span mathematical reasoning, code generation, knowledge reasoning, instruction following, and logical reasoning, showing that routing confidence provides a broadly effective signal for selective reasoning revision.

\paragraph{RoCo-PRM provides strong and transferable process supervision.}
RoCo-PRM achieves the best or tied-best performance in all 30 model--task settings across the five evaluation models, with 29 strict wins and one tie. Compared with the strongest PRM baseline, it improves average performance by 1.7, 1.5, and 1.5 points on OLMoE, DeepSeek, and Qwen1.5, respectively. The gains also transfer to dense models: RoCo-PRM improves over the strongest baseline by 1.4 points on Qwen2.5 and 1.6 points on Llama. These results demonstrate that routing-derived supervision remains effective across both MoE and dense target models.

\input{tables/main_results}

\section{Further Analysis}
\label{sec:further_analysis}

\subsection{Composite Routing Confidence Enables Better Revision-Point Selection}
\label{sec:analysis_routing_confidence}

We examine why combining the three routing statistics yields a stronger reasoning-guidance signal. We first ablate the components of routing confidence and then evaluate whether different formulations identify revision points with higher counterfactual corrective value.

\begin{figure}[t]
    \centering
    \includegraphics[width=0.9\linewidth]{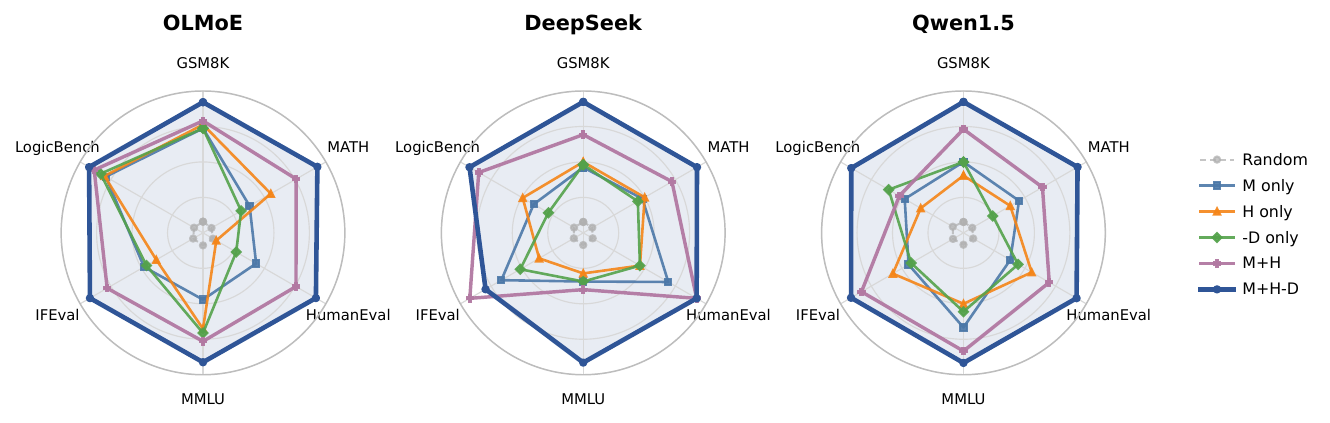}
    \vspace{-1em}
    \caption{
    \textbf{Routing-confidence ablation for RoCo-Rethink.}
Task axes use task-specific ranges. The full formulation is best or tied-best in 17 of 18 settings.}
    \label{fig:roco_ablation}
    \vspace{-1.5em}
\end{figure}

\paragraph{Component ablation.}
Figure~\ref{fig:roco_ablation} compares RoCo-Rethink with variants using only overall routing mass ($M$), head routing mass ($H$), or negative routing variation ($-D$), as well as their partial combination $M+H$. Random rethink is included as a reference to separate the effect of routing-guided revision from that of additional test-time computation alone.

Across the three MoE models, each individual routing statistic generally improves over random rethink, indicating that all three components provide useful signals for reasoning revision. Combining overall and head routing mass further strengthens performance: $M+H$ outperforms all single-component variants in 17 of 18 model--task settings, suggesting that global routing behavior and early-step routing provide complementary information. Adding negative routing variation yields the complete confidence score $M+H-D$, which achieves the best performance in 17 of 18 settings.

\paragraph{Revision-point quality.} 
To understand why the composite score leads to better rethink performance, we construct a Monte Carlo revision oracle on initially incorrect trajectories. For each candidate step $i$, we preserve the preceding prefix and sample $K$ regenerated suffixes using the same rethink policy. We define \begin{equation} V_i=\frac{1}{K}\sum_{k=1}^{K}\mathbb{I}[\hat{y}_i^{(k)}=y^\star], \qquad V^\star=\max_i V_i, \qquad R_g=V^\star-V_{i_g}, \end{equation}
where $V_i$ estimates the corrective value of revising from step $i$, $V^\star$ is the oracle revision value, and $R_g$ is the regret of a selector $g$ choosing step $i_g$. Lower regret indicates a higher-value revision point. 

\begin{wrapfigure}{R}{0.46\linewidth}
    \centering
    \vspace{-2em}
    \includegraphics[width=\linewidth]{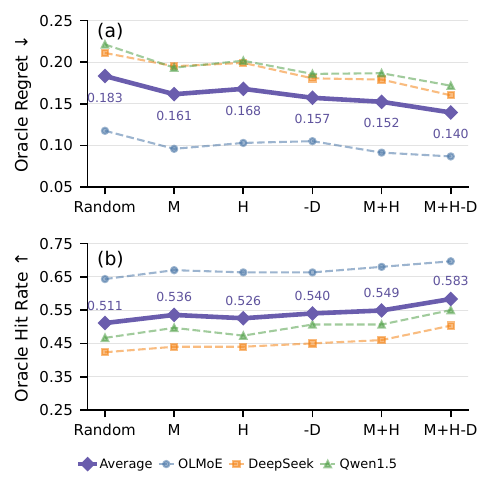}
    \vspace{-1.5em}
    \caption{
    \textbf{Monte Carlo revision-oracle analysis.}
    Oracle (a) regret ($\downarrow$) and (b) hit rate ($\uparrow$), averaged over three tasks. $M{+}H{-}D$ consistently achieves the best revision-point quality.
    }
    \label{fig:revision_oracle}
    \vspace{-3em}
\end{wrapfigure}

We also report oracle hit rate, the fraction of trajectories for which the selector chooses an oracle-optimal step. We evaluate GSM8K, HumanEval, and LogicBench across all three MoE models.

Figure~\ref{fig:revision_oracle} shows a consistent advantage for composite routing confidence. The full $M{+}H{-}D$ selector achieves the lowest oracle regret and the highest oracle hit rate in all nine model--task settings. Averaged across settings, its oracle regret is $0.140$, compared with $0.183$ for random selection and $0.152$ for the partial $M{+}H$ formulation, corresponding to a $23.5\%$ reduction relative to random selection. Its average oracle hit rate increases from $0.511$ under random selection to $0.583$, exceeding $M{+}H$ at $0.549$. While the strongest individual routing statistic varies across models and tasks, their full combination remains consistently strongest. These results suggest that the gains of composite routing confidence arise from its ability to identify revision points with higher counterfactual corrective value.

\subsection{RoCo-Rethink Uses Test-Time Rethink Budget More Effectively}
\label{sec:analysis_budget}

We analyze how effectively different rethink methods convert additional test-time budget into performance gains. We vary the maximum number of rethink rounds from 0 to 3 and report the mean gain over Vanilla on four benchmarks: GSM8K, MATH500, MMLU, and LogicBench. For each model--task pair, we compute the gain relative to Vanilla and average these gains across tasks.

\begin{figure}[H]
    \centering
    \includegraphics[width=0.9\linewidth]{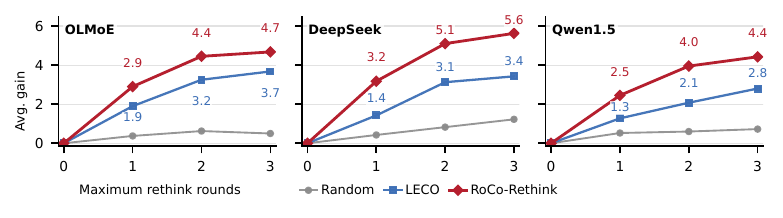}
    \vspace{-1em}
    \caption{
    \textbf{Test-time rethink-budget scaling.}
    RoCo-Rethink consistently yields larger gains than LECO and random rethink across all three MoE models.
    }
    \label{fig:rethink_budget}
    \vspace{-1em}
\end{figure}

Figure~\ref{fig:rethink_budget} shows that RoCo-Rethink consistently delivers larger gains than both LECO and random rethink under the same rethink-round budget. On OLMoE, the average gain of RoCo-Rethink increases from 2.9 at one rethink round to 4.7 at three rounds, compared with 1.9 to 3.7 for LECO. On DeepSeek, the corresponding gains are 3.2 to 5.6 for RoCo-Rethink and 1.4 to 3.4 for LECO. On Qwen1.5, RoCo-Rethink improves from 2.5 to 4.4, whereas LECO improves from 1.3 to 2.8. Random rethink yields only small and less stable gains across all models.

These results show that RoCo-Rethink not only achieves stronger final performance, but also converts additional rethink rounds into performance gains more effectively. Substantial improvements already emerge at low rethink budgets, indicating that routing confidence allocates extra test-time reasoning more effectively than logit-based confidence.

\subsection{RoCo-PRM Is More Sensitive to Teacher Quality than Backbone Scale}
\label{sec:prm_teacher_backbone}

We next examine whether RoCo-PRM is more strongly affected by the routing teacher or by the capacity of the standalone PRM. We vary the routing teacher while fixing the PRM backbone, and separately vary the PRM backbone while fixing the routing teacher.

\noindent
\begin{minipage}[t]{0.61\linewidth}
\vspace{0pt}

We first fix the PRM backbone to Qwen2.5-7B-Instruct and compare routing supervision generated by DeepSeek-V2-Lite-Chat and Qwen3-30B-A3B. Using \qwenthree~as the routing teacher yields an average improvement of 2.0 points over \deepseek~across the 30 model--task settings. The gain is consistent across all five target models, ranging from 1.4 to 2.7 points in average score. These results indicate that the quality of routing-derived supervision has a substantial influence on the effectiveness of the resulting process evaluator.

\end{minipage}
\hfill
\begin{minipage}[t]{0.36\linewidth}
\vspace{0pt}

\centering
\captionsetup{hypcap=false}
\captionof{table}{
\textbf{Teacher and backbone effects on RoCo-PRM.}
}
\label{tab:prm_teacher_backbone}
\vspace{-0.5em}

\footnotesize
\setlength{\tabcolsep}{3.5pt}
\renewcommand{\arraystretch}{0.95}
\begin{tabular}{lcc}
\toprule
Setting & Avg. & Gain \\
\midrule
DeepSeek teacher & 49.5 & -- \\
Qwen3 teacher & 51.5 & +2.0 \\
\midrule
Qwen2.5-1.5B backbone & 50.6 & -- \\
Qwen2.5-7B backbone & 51.5 & +0.9 \\
\bottomrule
\end{tabular}

\end{minipage}

We then fix Qwen3-30B-A3B as the routing teacher and compare Qwen2.5-1.5B-Instruct and Qwen2.5-7B-Instruct as PRM backbones. Scaling the backbone from 1.5B to 7B improves the average score by 0.9 points across the same 30 settings, with per-model gains ranging from 0.6 to 1.5 points. Notably, the 1.5B backbone already retains most of the performance achieved by the 7B model. Overall, RoCo-PRM is more sensitive to the quality of routing-derived supervision than to PRM backbone scale, suggesting that improving the routing teacher is more valuable than simply increasing the capacity of the downstream process evaluator.

\subsection{Routing-Derived Process Supervision Transfers Across Tasks}
\label{sec:prm_cross_task}

\noindent
\begin{minipage}[t]{0.52\linewidth}
\vspace{0pt}

Finally, we examine whether RoCo-PRM learns task-specific routing patterns or a more transferable notion of process quality. For each of four evaluation tasks (GSM8K, MATH500, MMLU, and HumanEval), we train a leave-one-task-out RoCo-PRM while excluding task-aligned routing supervision from the corresponding evaluation task. We then use the resulting PRM to guide three target models spanning both MoE and dense architectures.

\end{minipage}
\hfill
\begin{minipage}[t]{0.45\linewidth}
\vspace{0pt}
\centering

\includegraphics[width=\linewidth]{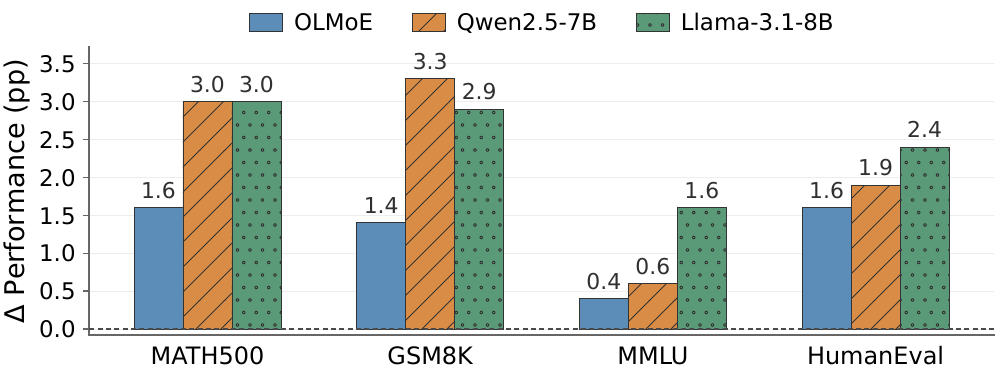}

\captionsetup{hypcap=false}
\captionof{figure}{
\textbf{Leave-one-task-out cross-task generalization of RoCo-PRM.}
}
\label{fig:loo_cross_task}

\end{minipage}

Figure~\ref{fig:loo_cross_task} shows consistent cross-task transfer. Despite excluding task-aligned routing supervision for the evaluation task, the leave-one-task-out RoCo-PRMs improve performance in all 12 target-model--task settings, with an average gain of $1.98$ points. The average gains reach $2.53$ points on both MATH500 and GSM8K and $1.97$ points on HumanEval. Transfer to MMLU is smaller, with an average gain of $0.87$ points, but remains positive across all three target models. These results indicate that routing-derived process supervision captures signals that generalize beyond the task from which the supervision is collected.

\section{Conclusion}

We introduced RoCo, a framework that turns native MoE routing into reasoning guidance and automatic process supervision. RoCo-Rethink uses routing confidence for targeted test-time revision, while RoCo-PRM distills routing-derived step scores into a standalone process evaluator that transfers across both MoE and dense models. Extensive experiments show consistent gains over strong test-time and PRM baselines, while further analyses link these gains to better revision-point selection, stronger routing teachers, and cross-task transfer of routing-derived supervision. Overall, our results establish expert routing as a practical, architecture-native source of process supervision and motivate broader exploration of internal computation signals for reasoning and learning.

\subsection*{AI use statement}

Generative AI tools were used to assist with manuscript drafting, language editing, and polishing; brainstorming and refinement of research hypotheses, experimental designs, and analyses; interpretation of experimental results; and coding, debugging, and visualization. All AI-assisted research outputs, including experimental designs, implementations, analyses, numerical results, and scientific claims, were reviewed and verified by the authors. Relevant literature and citations suggested with AI assistance were checked against the original sources. The authors take full responsibility for the final content of this work, including all text, claims, code, analyses, and artifacts produced with the aid of generative AI.

\subsection*{Reproducibility statement}

We provide the methodological details required to reproduce our results in the main paper and appendix. The appendix includes detailed definitions of routing features, dataset-specific supervision sources, decoding and rethink configurations, PRM training settings, baseline implementations, stopping criteria, and task-specific evaluation protocols. We also report results over multiple random seeds and describe the settings used for the Monte Carlo revision-oracle and ablation analyses. Code and scripts for data construction, routing extraction, training, and evaluation will be released to support reproducibility.



\bibliography{ref}
\bibliographystyle{iclr2027_conference}

\appendix

\section{Routing-Signal Definitions and Analysis Details}
\label{app:routing_features}

\subsection{Token-Level Routing Mass and Layer Normalization}
\label{app:token_routing}

For each generated token $t$ and MoE layer $\ell$, let $p_{t,\ell}(e)$ denote the router probability assigned to expert $e$. We define the selected-expert routing mass as
\begin{equation}
    r_{t,\ell}
    =
    \sum_{e\in\mathrm{TopK}_{t,\ell}} p_{t,\ell}(e),
\end{equation}
where $\mathrm{TopK}_{t,\ell}$ is the set of experts selected by the model's native top-$k$ routing rule. We use each model's native number of selected experts: $k=6$ for DeepSeek-V2-Lite-Chat, $k=8$ for OLMoE-1B-7B-0924-SFT, and $k=4$ for Qwen1.5-MoE-A2.7B-Chat.

The raw routing-mass scale can differ across MoE layers. The main RoCo experiments therefore use the layer-normalized routing variant. For each layer, we min--max normalize the token-level routing mass over the current generated trajectory:
\begin{equation}
    \bar r_{t,\ell}
    =
    \frac{r_{t,\ell}-\min_{t'}r_{t',\ell}}
    {\max_{t'}r_{t',\ell}-\min_{t'}r_{t',\ell}+\epsilon},
    \qquad \epsilon=10^{-8}.
\end{equation}
We then average the normalized mass across the $L$ MoE layers,
\begin{equation}
    r_t = \frac{1}{L}\sum_{\ell=1}^{L}\bar r_{t,\ell},
\end{equation}
obtaining one layer-aggregated routing signal for each generated token. Router probabilities are computed by applying softmax to the full router logits. For DeepSeek-V2-Lite-Chat, whose MoE gate directly returns selected expert indices and weights, we reconstruct the full router logits from the gate input and router weight before applying softmax.

\subsection{Step Segmentation and Routing Statistics}
\label{app:step_routing}

We segment a generated reasoning trajectory into newline-delimited non-empty reasoning steps. Token-to-step assignments are obtained from tokenizer offset mappings, which align each generated token with the character span of its corresponding reasoning step. Let $\mathcal{T}_i$ denote the token indices belonging to step $s_i$, and let $\mathcal{H}_i$ contain the first $h$ tokens of that step. We use $h=3$ in the main experiments. For steps shorter than $h$ tokens, all available tokens are used.

Using the layer-normalized token signal $r_t$, we define
\begin{equation}
    M_i = \frac{1}{|\mathcal{T}_i|}\sum_{t\in\mathcal{T}_i}r_t,
    \qquad
    H_i = \frac{1}{|\mathcal{H}_i|}\sum_{t\in\mathcal{H}_i}r_t,
    \qquad
    D_i = \operatorname{Std}_{t\in\mathcal{T}_i}(r_t).
\end{equation}
Here, $M_i$ captures the overall selected-expert mass of the step, $H_i$ emphasizes routing behavior near the beginning of the step, and $D_i$ measures within-step variation. The implementation uses the population standard deviation.

\subsection{Trajectory-Level Normalization and Routing Confidence}
\label{app:routing_confidence_details}

Because the ranges of $M$, $H$, and $D$ can differ within a trajectory, we apply a second min--max normalization across reasoning steps. For a generic step-level statistic $X_i\in\{M_i,H_i,D_i\}$,
\begin{equation}
    \widetilde X_i
    =
    \frac{X_i-\min_j X_j}
    {\max_j X_j-\min_j X_j+10^{-8}}.
\end{equation}
The routing-confidence score used by RoCo-Rethink is
\begin{equation}
    C_i = \widetilde M_i + \widetilde H_i - \widetilde D_i.
\end{equation}
Thus, the confidence is trajectory-relative rather than a globally calibrated probability. This is sufficient for RoCo-Rethink, which only requires a within-trajectory ordering to select the lowest-confidence step.

\subsection{Routing--Reliability Analysis}
\label{app:routing_reliability_details}

Section~\ref{sec:routing_reliability} evaluates whether the three routing statistics track independent estimates of step quality. For GSM8K and MATH500, we use two math-oriented process verifiers~\citep{zhang2025lessons,xiong2024implementation} and average their within-trajectory Spearman correlations for Figure~\ref{fig:routing_signal}(a). For code reasoning we use CodePRM~\citep{li2025codeprm} on HumanEval, and for knowledge reasoning we use VersaPRM~\citep{zeng2025versaprm} on MMLU. Table~\ref{tab:routing_spearman_full} reports the values used in the figure.

\begin{table}[t]
\caption{Within-trajectory Spearman correlation between routing statistics and verifier-assessed step quality. GSM8K and MATH500 values are averaged over the two math PRMs used in the main paper.}
\label{tab:routing_spearman_full}
\centering
\small
\begin{tabular}{llrrr}
\toprule
Model & Task & $M$ & $H$ & $D$ \\
\midrule
DeepSeek & GSM8K   & 0.106 & 0.062 & -0.100 \\
DeepSeek & MATH500 & 0.111 & 0.053 & -0.021 \\
DeepSeek & HumanEval    & 0.235 & 0.111 & -0.068 \\
DeepSeek & MMLU    & 0.289 & 0.236 & -0.169 \\
\midrule
OLMoE & GSM8K   & 0.249 & 0.252 & -0.113 \\
OLMoE & MATH500 & 0.028 & 0.073 & -0.090 \\
OLMoE & HumanEval    & 0.065 & 0.214 & -0.060 \\
OLMoE & MMLU    & 0.253 & 0.068 & -0.047 \\
\midrule
Qwen1.5 & GSM8K   & 0.089 & 0.072 & -0.019 \\
Qwen1.5 & MATH500 & 0.074 & 0.065 & -0.061 \\
Qwen1.5 & HumanEval    & 0.057 & 0.105 & -0.146 \\
Qwen1.5 & MMLU    & 0.365 & 0.213 & -0.015 \\
\bottomrule
\end{tabular}
\end{table}

For Figure~\ref{fig:routing_signal}(b), we compare lightweight bad-step predictors constructed from either routing-side or output-logit-side features. The routing predictor uses the three step-level statistics introduced in Section~\ref{sec:routing_reliability}: overall routing mass ($M$), head routing mass ($H$), and within-step routing variation ($D$). The logit predictor uses three corresponding output-side confidence features, providing a dimension-matched comparison between routing-side and output-side signals.

Both predictors use the same prediction pipeline consisting of median imputation, feature standardization, and logistic regression with balanced class weights and a maximum of 1,000 iterations. We evaluate them using five-fold GroupKFold cross-validation at the reasoning-trajectory level, ensuring that all steps from the same trajectory remain in the same fold. Within each trajectory, steps whose verifier scores fall at or below the 20th percentile are labeled as bad; trajectories with fewer than three valid steps are excluded. AUROC is computed from the out-of-fold predicted probability of the bad-step class over all valid steps. Across the 18 model--task--verifier settings shown in Figure~\ref{fig:routing_signal}(b), the routing predictor outperforms the logit predictor in 13 settings. Table~\ref{tab:routing_logit_auc} reports the complete verifier-specific results.

\begin{table}[t]
\caption{Grouped-CV bad-step AUROC for routing-based and logit-based predictors.}
\label{tab:routing_logit_auc}
\centering
\small
\setlength{\tabcolsep}{4pt}
\begin{tabular}{lllrr}
\toprule
Model & Task & Verifier & Logit & Routing \\
\midrule
DeepSeek & GSM8K & Llama-PRM & 0.542 & 0.534 \\
DeepSeek & GSM8K & Qwen-PRM  & 0.529 & 0.533 \\
DeepSeek & MATH500 & Llama-PRM & 0.576 & 0.583 \\
DeepSeek & MATH500 & Qwen-PRM  & 0.573 & 0.598 \\
DeepSeek & HumanEval & CodePRM & 0.577 & 0.592 \\
DeepSeek & MMLU & VersaPRM & 0.391 & 0.496 \\
OLMoE & GSM8K & Llama-PRM & 0.551 & 0.644 \\
OLMoE & GSM8K & Qwen-PRM & 0.560 & 0.592 \\
OLMoE & MATH500 & Llama-PRM & 0.622 & 0.644 \\
OLMoE & MATH500 & Qwen-PRM & 0.619 & 0.632 \\
OLMoE & HumanEval & CodePRM & 0.521 & 0.584 \\
OLMoE & HumanEval & VersaPRM & 0.420 & 0.498 \\
Qwen1.5 & GSM8K & Llama-PRM & 0.527 & 0.519 \\
Qwen1.5 & GSM8K & Qwen-PRM & 0.551 & 0.548 \\
Qwen1.5 & MATH500 & Llama-PRM & 0.568 & 0.584 \\
Qwen1.5 & MATH500 & Qwen-PRM & 0.557 & 0.537 \\
Qwen1.5 & HumanEval & CodePRM & 0.586 & 0.574 \\
Qwen1.5 & MMLU & VersaPRM & 0.461 & 0.642 \\
\bottomrule
\end{tabular}
\end{table}

\section{Experimental Setup Details}
\label{app:implementation}

\subsection{Model Configurations}

Table~\ref{tab:model_configs} summarizes the architectural configurations relevant to routing extraction. We use each MoE model's native top-$k$ routing rule rather than imposing a common $k$ across architectures.

\begin{table}[t]
\caption{Model configurations used in the experiments. Routing columns are shown for MoE models only.}
\label{tab:model_configs}
\centering
\small
\begin{tabular}{lrrrrl}
\toprule
Model & Layers & Experts & Top-$k$ & Hidden & Type \\
\midrule
DeepSeek-V2-Lite-Chat & 27 & 64 & 6 & 2048 & MoE \\
OLMoE-1B-7B-0924-SFT & 16 & 64 & 8 & 2048 & MoE \\
Qwen1.5-MoE-A2.7B-Chat & 24 & 60 & 4 & 2048 & MoE \\
Qwen3-30B-A3B & 48 & 128 & 8 & 2048 & MoE teacher \\
Qwen2.5-7B-Instruct & 28 & -- & -- & 3584 & Dense \\
Llama-3.1-8B-Instruct & 32 & -- & -- & 4096 & Dense \\
\bottomrule
\end{tabular}
\end{table}

\subsection{Benchmarks and Evaluation Protocols}
\label{app:benchmarks}

We evaluate six benchmarks spanning mathematical reasoning, knowledge reasoning, code generation, instruction following, and logical reasoning. Table~\ref{tab:benchmark_setup} summarizes the evaluation sets and primary metrics. All evaluation sets are disjoint from the data used to construct PRM supervision.

\begin{table}[t]
\centering
\small
\setlength{\tabcolsep}{4pt}
\caption{\textbf{Benchmark and evaluation setup.}}
\label{tab:benchmark_setup}
\begin{tabular}{lcl}
\toprule
Benchmark & Eval size & Metric \\
\midrule
GSM8K & 1,319 & Numerical answer accuracy \\
MATH500 & 500 & Numerical answer accuracy \\
MMLU & 14,042 & Multiple-choice accuracy \\
HumanEval & 164 & pass@1 \\
IFEval & 541 prompts / 834 instr. & Loose prompt accuracy \\
LogicBench & 1,520 QA pairs & BQA accuracy \\
\bottomrule
\end{tabular}
\end{table}

Detailed prompting, response parsing, and task-specific correctness criteria are provided in Appendix~\ref{app:prompts_eval}.

\subsection{Process-Supervision Data Sources}
\label{app:supervision_data}

RoCo-PRM supervision is constructed from task-aligned training corpora that are disjoint from the corresponding evaluation sets. For each source problem, the MoE teacher generates a reasoning trajectory, and each reasoning prefix is paired with its routing-confidence score to form a scalar-regression example. Table~\ref{tab:supervision_sources} summarizes the supervision sources used for each evaluation task and the resulting number of step-level training examples.

\begin{table}[t]
\centering
\small
\setlength{\tabcolsep}{4pt}
\caption{\textbf{Process-supervision data used to train RoCo-PRM.}
Counts denote final step-level regression examples used for training.}
\label{tab:supervision_sources}
\begin{tabular}{llr}
\toprule
Evaluation task & Supervision source & Step examples \\
\midrule
GSM8K      & GSM8K train & 155,391 \\
MATH500    & MATH train & 370,384 \\
MMLU       & MMLU train & 82,338 \\
HumanEval  & APPS train & 92,628 \\
IFEval     & IFEval supervision set & 48,122 \\
LogicBench & LogicBench augmentation & 104,846 \\
\bottomrule
\end{tabular}
\end{table}

The same routing-label construction procedure is applied across all supervision sources. To illustrate the resulting training data, Table~\ref{tab:prm_examples} shows representative prefix-level examples from GSM8K and LogicBench. Each reasoning prefix $s_{\le i}$ is paired with the trajectory-normalized routing statistics $\widetilde M_i$, $\widetilde H_i$, and $\widetilde D_i$, together with the corresponding regression target
$C_i=\widetilde M_i+\widetilde H_i-\widetilde D_i$.
For compactness, we display only the newly appended reasoning step for each prefix.

\begin{table}[t]
\centering
\small
\setlength{\tabcolsep}{3.5pt}
\caption{\textbf{Examples of routing-derived RoCo-PRM supervision.}
Each row corresponds to one prefix-level regression example $(x,s_{\le i},C_i)$. 
For compactness, only the newly appended reasoning step is shown. 
The routing statistics are normalized within the trajectory, and
$C_i=\widetilde M_i+\widetilde H_i-\widetilde D_i$.}
\label{tab:prm_examples}
\begin{tabular}{lp{0.53\linewidth}cccc}
\toprule
Task & Newly appended reasoning step
& $\widetilde M_i$
& $\widetilde H_i$
& $\widetilde D_i$
& $C_i$ \\
\midrule

\multirow{4}{*}{GSM8K}
& Let's count the legs:
& 1.000 & 1.000 & 1.000 & 1.000 \\

& Dogs: each dog has 4 legs; $2\times4=8$ legs.
& 0.384 & 0.407 & 0.363 & 0.427 \\

& Chicken: $1\times2=2$ legs.
& 0.000 & 0.000 & 0.145 & -0.145 \\

& Total legs: $8+2=10$.
& 0.049 & 0.240 & 0.243 & 0.046 \\

\midrule

\multirow{4}{*}{LogicBench}
& Step 1: Identify the logical rule or relation in the context.
& 1.000 & 1.000 & 1.000 & 1.000 \\

& The logical rule is that all cats love to play.
& 0.598 & 0.480 & 0.090 & 0.987 \\

& Step 2: Apply the rule to the facts in the question.
& 0.378 & 0.157 & 0.788 & -0.252 \\

& Since Tom is a cat, the rule applies to Tom.
& 0.292 & 0.124 & 0.046 & 0.371 \\

\bottomrule
\end{tabular}
\end{table}

\section{RoCo-Rethink Implementation Details}
\label{app:rethink_details}

\subsection{Post-Hoc Revision Procedure}

RoCo-Rethink operates post hoc. The model first produces a complete reasoning trajectory, after which we replay the trajectory to extract router probabilities and compute routing confidence for every reasoning step. The lowest-confidence step is selected as the revision point. We preserve all reasoning steps preceding that point and regenerate the suffix beginning from the selected step. After each regeneration, routing confidence is recomputed over the updated trajectory and a new revision point is selected.

For tasks with an extractable final answer, the implementation stops early
when two successive extracted answers are identical, subject to the configured
minimum number of rethink rounds. The reported answer is obtained by majority
vote over the initial answer and all rethink answers; ties are resolved in
favor of the most recent answer. For HumanEval and IFEval, where no compact
answer string is available, convergence is determined from the generated
completion or response, respectively, and the latest output is used for
evaluation. For random rethink, the revision step is drawn uniformly from the available reasoning steps.

In the main experiments, the initial response is generated greedily and rethink suffixes use mild sampling ($T=0.3$, top-$p=0.95$), with at most three rethink rounds.

\subsection{Test-Time Baselines}

\paragraph{Self-Consistency.}
For non-code benchmarks, the implementation samples eight independent completions by default using temperature $0.7$ and top-$p=0.95$, and aggregates extracted answers by majority voting. For HumanEval, eight independent code completions are generated and evaluated with the official functional-correctness harness. To keep the main comparison aligned with the rest of the table, we report the official pass@1 estimate; pass@8 is logged as an auxiliary diagnostic and is not used as the main-table score.

\paragraph{LECO.}
LECO uses output-logit-derived step confidence to select a low-confidence revision point and otherwise follows the same selective suffix-regeneration protocol as RoCo-Rethink. To ensure a controlled comparison, we construct its confidence score using the same three-part decomposition as RoCo: overall token confidence, head token confidence, and within-step confidence variation. Token-level confidence is defined as the probability assigned to the realized token under teacher forcing. The three step-level statistics are normalized within each trajectory and combined using the same formulation,
\[
C_i^{\mathrm{LECO}}
=
\widetilde M_i^{\mathrm{logit}}
+
\widetilde H_i^{\mathrm{logit}}
-
\widetilde D_i^{\mathrm{logit}}.
\]
Thus, LECO and RoCo-Rethink differ only in the source of the confidence signal: output logits versus native MoE routing.

\paragraph{Random rethink.}
Random rethink uses the same suffix-regeneration machinery but selects the revision point uniformly at random from the available steps. This baseline separates gains from confidence-guided revision from gains due solely to additional generation.

\section{Additional RoCo-Rethink Results}
\label{app:rethink_additional}

\subsection{Routing-Component Ablation}
\label{app:routing_ablation}

Table~\ref{tab:routing_component_ablation} reports the complete routing-component ablation underlying Figure~\ref{fig:roco_ablation}. We compare random rethink with variants using overall routing mass ($M$), head routing mass ($H$), negative routing variation ($-D$), their partial combination ($M+H$), and the complete routing-confidence formulation ($M+H-D$). Across the three MoE models and six tasks, $M+H$ outperforms all individual routing statistics in 17 of 18 settings, while $M+H-D$ achieves the best or tied-best performance in 17 of 18 settings.

\begin{table}[t]
\centering
\small
\setlength{\tabcolsep}{3.5pt}
\renewcommand{\arraystretch}{0.92}
\caption{\textbf{Full routing-component ablation for RoCo-Rethink.}
Results are reported as means over five random seeds; standard deviations are omitted for readability. Bold and underline indicate the best and second-best results within each model--task setting, respectively.}
\label{tab:routing_component_ablation}
\resizebox{\linewidth}{!}{
\begin{tabular}{llcccccc}
\toprule
Model & Method & GSM8K & MATH500 & HumanEval & MMLU & IFEval & LogicBench \\
\midrule

\multirow{6}{*}{OLMoE}
& Random
& 48.4 & 9.5 & 29.2 & 47.2 & 26.3 & 48.5 \\
& $M$
& 50.9 & 10.4 & 30.5 & 48.5 & 28.3 & 55.4 \\
& $H$
& 51.0 & 10.9 & 29.3 & 49.2 & 27.8 & 55.5 \\
& $-D$
& 50.9 & 10.2 & 29.9 & 49.3 & 28.2 & 55.8 \\
& $M+H$
& \underline{51.1} & \underline{11.5} & \underline{31.7}
& \underline{49.5} & \underline{29.8} & \underline{56.3} \\
& $M+H-D$
& \textbf{51.6} & \textbf{12.0} & \textbf{32.3}
& \textbf{50.0} & \textbf{30.5} & \textbf{56.7} \\
\midrule

\multirow{6}{*}{DeepSeek}
& Random
& 70.4 & 22.4 & 47.2 & 49.6 & 30.0 & 66.7 \\
& $M$
& 72.4 & 24.6 & \underline{48.8} & 50.5 & 32.3 & 68.9 \\
& $H$
& 72.6 & 24.7 & 48.2 & 50.3 & 31.1 & 69.5 \\
& $-D$
& 72.5 & 24.4 & 48.2 & 50.5 & 31.7 & 68.1 \\
& $M+H$
& \underline{73.6} & \underline{25.9} & \textbf{49.4}
& \underline{50.7} & \textbf{33.3} & \underline{71.9} \\
& $M+H-D$
& \textbf{74.8} & \textbf{27.0} & \textbf{49.4}
& \textbf{52.5} & \underline{32.8} & \textbf{72.4} \\
\midrule

\multirow{6}{*}{Qwen1.5}
& Random
& 50.2 & 17.2 & 18.5 & 55.7 & 26.8 & 65.1 \\
& $M$
& 52.4 & 18.8 & 20.4 & 57.8 & 28.5 & 66.9 \\
& $H$
& 51.9 & 18.5 & 21.5 & 57.2 & 29.1 & 66.3 \\
& $-D$
& 52.4 & 17.9 & 20.8 & 57.4 & 28.4 & \underline{67.5} \\
& $M+H$
& \underline{53.6} & \underline{19.6} & \underline{22.4}
& \underline{58.4} & \underline{30.3} & 67.1 \\
& $M+H-D$
& \textbf{54.6} & \textbf{20.8} & \textbf{23.8}
& \textbf{58.7} & \textbf{30.7} & \textbf{68.9} \\
\bottomrule
\end{tabular}}
\end{table}

\subsection{Monte Carlo Revision Oracle}
\label{app:revision_oracle}

We provide the complete per-setting results for the Monte Carlo revision-oracle analysis in Section~\ref{sec:further_analysis}. For each initially incorrect trajectory and each candidate revision step $i$, we preserve the prefix before step $i$ and sample $K$ regenerated suffixes using the same rethink policy. The revision value and selector regret are
\begin{equation}
V_i = \frac{1}{K}\sum_{k=1}^{K}\mathbb{I}[\hat y_i^{(k)}=y^\star],
\qquad
V^\star = \max_i V_i,
\qquad
R_g = V^\star - V_{i_g},
\end{equation}
where $i_g$ denotes the step selected by method $g$. Lower regret indicates that the selected step has higher counterfactual corrective value. We additionally report oracle hit rate, defined as the fraction of trajectories for which the selected step attains $V^\star$.

Table~\ref{tab:revision_oracle_full} reports the complete results across the nine model--task settings underlying Figure~\ref{fig:revision_oracle}. The full $M{+}H{-}D$ selector achieves the lowest regret and highest hit rate in all nine settings. Averaged across settings, it reduces oracle regret from $0.183$ under random selection to $0.140$, while increasing oracle hit rate from $0.511$ to $0.583$.

\begin{table*}[t]
\centering
\scriptsize
\setlength{\tabcolsep}{4pt}
\caption{
\textbf{Full Monte Carlo revision-oracle results.}
Regret measures the gap between the oracle revision value and the value of the selected step, while hit rate measures how often the selected step is oracle-optimal. Lower regret and higher hit rate are better.
}
\label{tab:revision_oracle_full}
\begin{tabular}{llrrrrrrrrrrrr}
\toprule
& &
\multicolumn{2}{c}{Random} &
\multicolumn{2}{c}{$M$} &
\multicolumn{2}{c}{$H$} &
\multicolumn{2}{c}{$-D$} &
\multicolumn{2}{c}{$M+H$} &
\multicolumn{2}{c}{$M+H-D$} \\
\cmidrule(lr){3-4}
\cmidrule(lr){5-6}
\cmidrule(lr){7-8}
\cmidrule(lr){9-10}
\cmidrule(lr){11-12}
\cmidrule(lr){13-14}
Task & Model &
Regret & Hit &
Regret & Hit &
Regret & Hit &
Regret & Hit &
Regret & Hit &
Regret & Hit \\
\midrule

LogicBench
& OLMoE
& .125 & .56
& .101 & .60
& .110 & .60
& .125 & .59
& .099 & .60
& \textbf{.094} & \textbf{.63} \\

& DeepSeek
& .249 & .31
& .230 & .32
& .239 & .32
& .199 & .35
& .218 & .34
& \textbf{.186} & \textbf{.42} \\

& Qwen1.5
& .211 & .41
& .166 & .45
& .176 & .42
& .184 & .42
& .158 & .46
& \textbf{.156} & \textbf{.47} \\

\midrule
GSM8K
& OLMoE
& .135 & .62
& .100 & .64
& .108 & .64
& .111 & .63
& .096 & .67
& \textbf{.091} & \textbf{.68} \\

& DeepSeek
& .256 & .38
& .236 & .40
& .235 & .40
& .224 & .40
& .220 & .41
& \textbf{.206} & \textbf{.44} \\

& Qwen1.5
& .338 & .24
& .319 & .26
& .321 & .25
& .288 & .31
& .316 & .26
& \textbf{.275} & \textbf{.37} \\

\midrule
HumanEval
& OLMoE
& .093 & .75
& .086 & .77
& .091 & .75
& .079 & .77
& .079 & .77
& \textbf{.075} & \textbf{.78} \\

& DeepSeek
& .128 & .58
& .119 & .60
& .124 & .60
& .119 & .60
& .100 & .63
& \textbf{.088} & \textbf{.65} \\

& Qwen1.5
& .115 & .75
& .095 & .78
& .108 & .75
& .086 & .79
& .086 & .80
& \textbf{.084} & \textbf{.81} \\

\midrule
\multicolumn{2}{l}{Average}
& .183 & .511
& .161 & .536
& .168 & .526
& .157 & .540
& .152 & .549
& \textbf{.140} & \textbf{.583} \\
\bottomrule
\end{tabular}
\end{table*}

\subsection{Rethink-Budget Scaling}
\label{app:budget_scaling}

Table~\ref{tab:budget_scaling_full} reports the four-task mean gain over Vanilla for the budget-scaling analysis in the main paper. The gain is computed separately within each model--task pair and then averaged across GSM8K, MATH500, MMLU, and LogicBench.

\begin{table}[t]
\caption{Mean score gain over Vanilla as the maximum rethink-round budget $B$ increases.}
\label{tab:budget_scaling_full}
\centering
\small
\begin{tabular}{llrrr}
\toprule
Model & Method & $B=1$ & $B=2$ & $B=3$ \\
\midrule
OLMoE & Random & 0.38 & 0.62 & 0.50 \\
OLMoE & LECO & 1.90 & 3.25 & 3.68 \\
OLMoE & RoCo-Rethink & 2.90 & 4.45 & 4.68 \\
\midrule
DeepSeek & Random & 0.43 & 0.83 & 1.23 \\
DeepSeek & LECO & 1.43 & 3.13 & 3.43 \\
DeepSeek & RoCo-Rethink & 3.18 & 5.10 & 5.63 \\
\midrule
Qwen1.5 & Random & 0.53 & 0.60 & 0.73 \\
Qwen1.5 & LECO & 1.28 & 2.08 & 2.80 \\
Qwen1.5 & RoCo-Rethink & 2.45 & 3.95 & 4.43 \\
\bottomrule
\end{tabular}
\end{table}




\section{RoCo-PRM Training Details}
\label{app:prm_training}

\subsection{Routing-Derived Training Examples}
\label{app:routing_training_examples}

Each teacher-generated trajectory is expanded into one regression example per reasoning step, including the final answer step. For step $s_i$, the RoCo-PRM input consists of the original problem $x$ and the reasoning prefix $s_{\le i}$. We apply no additional step-level filtering beyond removing empty steps and samples with invalid or missing routing features.

For the layer-normalized supervision used in our main experiments, the routing statistics of all valid steps in the trajectory are min--max normalized to obtain $\widetilde M_i$, $\widetilde H_i$, and $\widetilde D_i$. The regression target is then
\begin{equation}
    C_i
    =
    \widetilde M_i
    +
    \widetilde H_i
    -
    \widetilde D_i.
\end{equation}
Each training example therefore takes the form $(x,s_{\le i},C_i)$. No maximum or minimum number of reasoning steps is imposed during dataset construction, and reasoning prefixes are not truncated at this stage; sequence truncation is applied only during tokenization for PRM training.

\subsection{PRM Architecture and Optimization}
\label{app:prm_architecture}

We instantiate RoCo-PRM from Qwen2.5-7B-Instruct with a scalar sequence-regression head that predicts the routing-derived confidence target $C_i$ from the problem and reasoning prefix $(x,s_{\le i})$. The training objective is mean squared error (MSE):
\begin{equation}
    \mathcal{L}_{\mathrm{PRM}}
    =
    \frac{1}{N}
    \sum_{n=1}^{N}
    \left(
        f_\theta(x^{(n)},s_{\le i}^{(n)})
        -
        C_i^{(n)}
    \right)^2.
\end{equation}

We fine-tune the model with LoRA while keeping the scalar regression head trainable. LoRA adapters are applied to the attention projections
(\texttt{q\_proj}, \texttt{k\_proj}, \texttt{v\_proj}, and \texttt{o\_proj})
and the MLP projections
(\texttt{gate\_proj}, \texttt{up\_proj}, and \texttt{down\_proj}).
Unless otherwise stated, we train for one epoch and use the final checkpoint; we do not apply early stopping or validation-based checkpoint selection. Inputs longer than 1,024 tokens are truncated during training tokenization, while no truncation is applied during supervision-data construction.

\begin{table}[t]
\centering
\small
\setlength{\tabcolsep}{5pt}
\caption{\textbf{Default RoCo-PRM training configuration.}}
\label{tab:prm_training_config}
\begin{tabular}{ll}
\toprule
Configuration & Value \\
\midrule
Backbone & Qwen2.5-7B-Instruct \\
Objective & Scalar regression (MSE) \\
Optimizer & AdamW (\texttt{adamw\_torch}) \\
Weight decay & 0 \\
LoRA rank & 16 \\
LoRA $\alpha$ & 32 \\
LoRA dropout & 0.05 \\
LoRA target modules
& \texttt{q/k/v/o\_proj}, \texttt{gate/up/down\_proj} \\
Learning rate & $2\times10^{-4}$ \\
Scheduler & Linear \\
Warmup & 0 \\
Epochs & 1 \\
Per-device batch size & 1 \\
Gradient accumulation & 32 \\
GPUs & 4 \\
Global effective batch size & 128 \\
Maximum sequence length & 1,024 \\
Precision & BF16 \\
Gradient checkpointing & Yes \\
Validation split & 5\% \\
Checkpoint selection & Final checkpoint \\
\bottomrule
\end{tabular}
\end{table}

Table~\ref{tab:prm_training_config} summarizes the default training configuration. We reserve 5\% of the training examples as a held-out validation split and report MSE, MAE, and Spearman correlation between predicted and routing-derived confidence scores on this split.

\subsection{PRM-Guided Rethink}

At test time, the trained PRM scores each reasoning prefix of a target-model trajectory. The scores are min--max normalized within the trajectory, and the lowest-scoring step is selected as the revision point. The target model then regenerates the suffix from that point using the same selective-rethink protocol as RoCo-Rethink. Because routing is used only to construct the teacher labels, the resulting PRM can guide dense models without access to target-model routing information.

\section{PRM Baseline Details}
\label{app:prm_baselines}

To isolate the effect of supervision construction, MC-PRM, CPMI-PRM, and RoCo-PRM use the same Qwen2.5-7B-Instruct regression backbone and the same training recipe unless otherwise stated. All three methods therefore differ primarily in how their step-level supervision targets are constructed.

\subsection{MC-PRM}

MC-PRM estimates the quality of each reasoning prefix through Monte Carlo continuation. Given a prefix $s_{\le i}$, we append it to the original problem prompt and sample $K=8$ continuation suffixes using Qwen3-30B-A3B with temperature $0.7$ and top-$p=0.95$. The maximum generation length is task dependent: 256 tokens for GSM8K, 512 for MATH500, 128 for MMLU, and 256 for LogicBench.

Let $\hat{y}_{i}^{(k)}$ denote the outcome obtained from the $k$-th rollout from prefix $s_{\le i}$. The MC supervision target is
\begin{equation}
    y_i^{\mathrm{MC}}
    =
    \frac{1}{K}
    \sum_{k=1}^{K}
    \mathbb{I}
    \left[
        \hat{y}_{i}^{(k)} \text{ is correct}
    \right],
    \qquad K=8.
\end{equation}

Correctness is determined using the task-specific evaluator. For GSM8K and MATH500, we extract and normalize the final numerical answer. For MMLU, we extract the final A--D option, while LogicBench BQA uses the final yes/no answer. For HumanEval, each code prefix is completed to a full program and evaluated with the official HumanEval execution harness; the target is the fraction of rollouts that pass the unit tests. For IFEval, each rollout is evaluated with the official instruction-following checker, and strict prompt-level compliance is used to construct the MC-PRM supervision target. Loose prompt-level accuracy is retained as the evaluation metric reported in the main experiments.

The resulting target lies in $[0,1]$ and is used to train the same Qwen2.5-7B scalar-regression PRM architecture as RoCo-PRM.

\subsection{CPMI-PRM}

CPMI-PRM measures the incremental evidence contributed by the current reasoning step toward a correct target relative to plausible incorrect alternatives. For a candidate target $y$, define
\begin{equation}
    \Delta_i(y)
    =
    \ell(y \mid x,s_{\le i})
    -
    \ell(y \mid x,s_{<i}),
\end{equation}
where $\ell$ denotes the mean token log-probability of the target continuation. We compute these values using Qwen3-30B-A3B with vLLM prompt log-probabilities. The use of mean rather than summed token log-probability reduces sensitivity to target length.

For a positive target $y^+$ and a negative set $\mathcal{N}$, we define
\begin{equation}
    s_i^{\mathrm{CPMI}}
    =
    \Delta_i(y^+)
    -
    \frac{1}{|\mathcal{N}|}
    \sum_{y^- \in \mathcal{N}}
    \Delta_i(y^-),
\end{equation}
and convert the score to a bounded regression target,
\begin{equation}
    y_i^{\mathrm{CPMI}}
    =
    \sigma\!\left(s_i^{\mathrm{CPMI}}\right),
\end{equation}
corresponding to a scaling factor $\gamma=1$.

\paragraph{Discrete-answer tasks.}
For GSM8K and MATH500, negatives are numerical perturbations of the gold answer. We prioritize offsets of $+1$, $-1$, $+2$, and $-2$, followed by sign reversal, multiplicative perturbations ($1.1\times$ and $0.9\times$), and simple constants such as $0$, $1$, and $-1$, retaining at most four negatives. For MMLU, the three non-gold answer choices form the negative set. For LogicBench BQA, the opposite Boolean answer is used as the negative.

\paragraph{HumanEval.}
For code generation, the CPMI target is defined over executable completions rather than short answer strings. A passing completion serves as the positive target, while negatives are execution-failing completions drawn from model-sampled failures, previously generated incorrect completions for the same problem, and syntactically valid programs that fail the unit tests. Because code completions can vary substantially in length, we use mean token log-probability and construct positive and negative targets with approximately matched lengths whenever possible. The same incremental-evidence formulation is then applied to these completion-level targets.

\paragraph{IFEval.}
For instruction following, we use a constraint-aware completion-level CPMI construction. Candidate responses are generated from the same teacher-model family and filtered with the official IFEval checker. Responses satisfying the constraints are used as positive targets, while failed responses form the negative pool. We retain approximately one to three positives and three to eight negatives per prompt when available, with preference for negatives that are fluent but violate only a small subset of constraints. We further stratify IFEval negatives by failure severity. \emph{Hard} negatives violate only one constraint, \emph{medium} negatives satisfy part of the instruction set but still fail the final checker, and \emph{easy} negatives exhibit more obvious non-compliance. CPMI construction prioritizes hard and medium negatives to reduce the risk that the PRM merely learns to distinguish generally poor responses from instruction-compliant ones. For a response prefix $s_{\le i}$, candidate scoring is performed only on the remaining continuation rather than replaying the full response, avoiding duplication between the observed prefix and the candidate completion. Positive candidates are defined primarily using strict IFEval compliance; loose compliance is used only when insufficient strict positives are available. Candidate metadata retains strict and loose pass indicators together with per-instruction failure information.

Across all tasks, CPMI-derived scalar targets are used to train the same Qwen2.5-7B-Instruct regression PRM with the same LoRA and optimization configuration as RoCo-PRM and MC-PRM.

\section{Additional RoCo-PRM Analysis}
\label{app:prm_additional}

\subsection{Teacher and Backbone Ablations}
\label{app:teacher_backbone}

We provide the complete results underlying the teacher- and backbone-ablation analysis in the main paper. To isolate the effect of routing-teacher quality, we first fix the PRM backbone to Qwen2.5-7B-Instruct and compare supervision generated by DeepSeek-V2-Lite-Chat and Qwen3-30B-A3B. As shown in Table~\ref{tab:teacher_ablation_full}, using Qwen3 as the routing teacher improves the average score by 2.0 points across the 30 target-model--task settings. The improvement is consistent across all five target models, with average gains ranging from 1.4 to 2.7 points.

\begin{table}[t]
\centering
\small
\setlength{\tabcolsep}{3.5pt}
\caption{\textbf{Effect of routing-teacher quality on RoCo-PRM.}
The PRM backbone is fixed to Qwen2.5-7B-Instruct. Results are reported as mean scores; standard deviations are omitted for readability.}
\label{tab:teacher_ablation_full}
\resizebox{\linewidth}{!}{
\begin{tabular}{llccccccc}
\toprule
Target & Teacher & GSM8K & MATH500 & HumanEval & MMLU & IFEval & LogicBench & Avg. \\
\midrule
\multirow{2}{*}{OLMoE}
& DeepSeek & 50.1 & 10.8 & 30.4 & 48.5 & 28.3 & 53.5 & 36.9 \\
& Qwen3    & \textbf{53.0} & \textbf{13.4} & \textbf{32.9} & \textbf{51.3} & \textbf{29.4} & \textbf{57.8} & \textbf{39.6} \\
\midrule
\multirow{2}{*}{DeepSeek}
& DeepSeek & 72.2 & 25.5 & 47.8 & 51.2 & 31.5 & 69.9 & 49.7 \\
& Qwen3    & \textbf{73.8} & \textbf{26.7} & \textbf{48.8} & \textbf{51.9} & \textbf{33.1} & \textbf{72.8} & \textbf{51.2} \\
\midrule
\multirow{2}{*}{Qwen1.5}
& DeepSeek & 52.9 & 19.5 & 20.1 & 56.4 & 28.9 & 67.3 & 40.9 \\
& Qwen3    & \textbf{54.3} & \textbf{21.4} & \textbf{23.4} & \textbf{58.5} & \textbf{30.9} & \textbf{70.9} & \textbf{43.2} \\
\midrule
\multirow{2}{*}{Llama}
& DeepSeek & 79.2 & 32.0 & 51.2 & 61.7 & 49.9 & 69.4 & 57.2 \\
& Qwen3    & \textbf{81.4} & \textbf{34.2} & \textbf{51.8} & \textbf{63.4} & \textbf{51.3} & \textbf{71.5} & \textbf{58.9} \\
\midrule
\multirow{2}{*}{Qwen2.5}
& DeepSeek & 83.4 & 53.6 & 52.1 & 70.3 & 46.7 & 71.0 & 62.9 \\
& Qwen3    & \textbf{85.0} & \textbf{55.4} & \textbf{54.3} & \textbf{70.7} & \textbf{48.1} & \textbf{72.1} & \textbf{64.3} \\
\bottomrule
\end{tabular}}
\end{table}

We next fix Qwen3-30B-A3B as the routing teacher and vary the standalone PRM backbone between Qwen2.5-1.5B-Instruct and Qwen2.5-7B-Instruct. Table~\ref{tab:backbone_ablation_full} shows that increasing the PRM backbone size yields a substantially smaller average gain of 0.90 points across the same 30 settings, with per-model improvements ranging from 0.55 to 1.53 points.

\begin{table}[t]
\centering
\small
\setlength{\tabcolsep}{3.5pt}
\caption{\textbf{Effect of PRM backbone scale.}
The routing teacher is fixed to Qwen3-30B-A3B. Results are reported as mean scores; standard deviations are omitted for readability.}
\label{tab:backbone_ablation_full}
\resizebox{\linewidth}{!}{
\begin{tabular}{llccccccc}
\toprule
Target & PRM backbone & GSM8K & MATH500 & HumanEval & MMLU & IFEval & LogicBench & Avg. \\
\midrule
\multirow{2}{*}{OLMoE}
& Qwen2.5-1.5B & 51.3 & 11.5 & 31.5 & 49.2 & 28.9 & 56.2 & 38.1 \\
& Qwen2.5-7B   & \textbf{53.0} & \textbf{13.4} & \textbf{32.9} & \textbf{51.3} & \textbf{29.4} & \textbf{57.8} & \textbf{39.6} \\
\midrule
\multirow{2}{*}{DeepSeek}
& Qwen2.5-1.5B & 73.0 & 26.6 & \textbf{48.8} & 51.8 & 31.9 & 71.7 & 50.6 \\
& Qwen2.5-7B   & \textbf{73.8} & \textbf{26.7} & \textbf{48.8} & \textbf{51.9} & \textbf{33.1} & \textbf{72.8} & \textbf{51.2} \\
\midrule
\multirow{2}{*}{Qwen1.5}
& Qwen2.5-1.5B & 54.1 & 19.9 & 22.6 & 57.7 & 30.4 & 67.9 & 42.1 \\
& Qwen2.5-7B   & \textbf{54.3} & \textbf{21.4} & \textbf{23.4} & \textbf{58.5} & \textbf{30.9} & \textbf{70.9} & \textbf{43.2} \\
\midrule
\multirow{2}{*}{Llama}
& Qwen2.5-1.5B & 80.5 & 33.9 & 50.7 & 62.6 & 50.8 & 70.8 & 58.2 \\
& Qwen2.5-7B   & \textbf{81.4} & \textbf{34.2} & \textbf{51.8} & \textbf{63.4} & \textbf{51.3} & \textbf{71.5} & \textbf{58.9} \\
\midrule
\multirow{2}{*}{Qwen2.5}
& Qwen2.5-1.5B & 84.2 & 54.7 & 53.5 & \textbf{70.7} & 47.6 & 71.5 & 63.7 \\
& Qwen2.5-7B   & \textbf{85.0} & \textbf{55.4} & \textbf{54.3} & \textbf{70.7} & \textbf{48.1} & \textbf{72.1} & \textbf{64.3} \\
\bottomrule
\end{tabular}}
\end{table}

Overall, these results consistently indicate that RoCo-PRM is more sensitive to the quality of routing-derived supervision than to the capacity of the downstream PRM. This suggests that improving the routing teacher is more effective than simply scaling the process evaluator.

\subsection{Leave-One-Task-Out Cross-Task Generalization}
\label{app:loo_prm}

We further evaluate whether routing-derived process supervision transfers across tasks rather than merely capturing task-specific routing patterns. We construct four leave-one-task-out (LOO) RoCo-PRMs corresponding to GSM8K, MATH500, MMLU, and HumanEval. For each evaluation task, we exclude its task-aligned routing supervision from PRM training and train Qwen2.5-7B-Instruct on the remaining routing-labeled sources. The resulting PRM therefore receives no task-aligned routing supervision from the corresponding evaluation task.

The underlying supervision pool is constructed from GSM8K, MATH, MMLU, APPS, MBPP, and IFEval. As in the main RoCo-PRM experiments, each reasoning prefix is paired with its routing-derived confidence target, and the appropriate task-specific prompt format is retained during PRM training. All LOO PRMs use the same scalar-regression architecture and optimization recipe as the default RoCo-PRM. We train for one epoch in BF16 with DeepSpeed Stage 2, a maximum sequence length of 1,024, learning rate $2\times10^{-4}$, per-device batch size 1, gradient accumulation 32, and four training processes.

At evaluation time, each LOO PRM is used to guide the corresponding held-out task on three target models: OLMoE, Qwen2.5-7B-Instruct, and Llama-3.1-8B-Instruct. Table~\ref{tab:loo_prm_full} reports the complete results underlying Figure~\ref{fig:loo_cross_task}.

\begin{table}[t]
\centering
\small
\setlength{\tabcolsep}{4pt}
\caption{
\textbf{Full leave-one-task-out cross-task generalization results.}
Each RoCo-PRM is trained without task-aligned routing supervision from the corresponding evaluation task. Values show target-model performance before and after LOO RoCo-PRM guidance.
}
\label{tab:loo_prm_full}
\begin{tabular}{lcccc}
\toprule
Target model & MATH500 & GSM8K & MMLU & HumanEval \\
\midrule
OLMoE
& 10.2$\rightarrow$11.8
& 50.6$\rightarrow$52.0
& 49.2$\rightarrow$49.6
& 29.9$\rightarrow$31.5 \\

Qwen2.5-7B
& 50.0$\rightarrow$53.0
& 81.2$\rightarrow$84.5
& 70.0$\rightarrow$70.6
& 49.3$\rightarrow$51.2 \\

Llama-3.1-8B
& 29.8$\rightarrow$32.8
& 76.3$\rightarrow$79.2
& 60.5$\rightarrow$62.1
& 47.6$\rightarrow$50.0 \\
\midrule
Avg. gain
& +2.53
& +2.53
& +0.87
& +1.97 \\
\bottomrule
\end{tabular}
\end{table}

The LOO RoCo-PRMs improve performance in all 12 target-model--task settings, with an average gain of $1.98$ points. The strongest average transfer gains are observed on MATH500 and GSM8K, both at $+2.53$ points, followed by HumanEval at $+1.97$ points. MMLU exhibits a smaller average gain of $+0.87$ points, but the improvement remains positive for all three target models. These results provide additional evidence that routing-derived process supervision captures transferable process-level information rather than being limited to task-specific routing patterns.

\section{Prompting, Parsing, and Evaluation Details}
\label{app:prompts_eval}

\subsection{Task-Specific Prompting}

For GSM8K and MATH500, the user message is the original problem statement provided by the benchmark. For chat models, we wrap this message using the model tokenizer's native chat template with generation enabled. When no chat template is available, we use the fallback prompt:
\begin{quote}
\ttfamily
Question: \{question\}

Answer: Let's think step by step.
\end{quote}
The same prompting policy is used when reconstructing problem--prefix inputs for routing extraction and rethink inference.

For MMLU, the user message contains the benchmark-formatted multiple-choice question together with four labeled answer options and an instruction to return the corresponding option letter. The generated response is evaluated by extracting a prediction from $\{A,B,C,D\}$.

For LogicBench BQA, we use the following task-specific prompt~\citep{parmar2024logicbench}:
\begin{quote}
\ttfamily
Answer the following logic question. Think briefly, then give the final answer as yes or no.

Context:\\
\{context\}

Question:\\
\{question\}

Final answer:
\end{quote}
This task-specific prompt is then wrapped with the model's native chat template for chat-based models.

For HumanEval, we use the canonical benchmark continuation prompt consisting of the original function signature and docstring. We do not apply a chat template; generated code is appended directly to the benchmark prompt. Generation uses code-oriented stop sequences to avoid unrelated trailing text or subsequent function/class definitions.

For IFEval, the original instruction-following prompt is used directly as the user message and wrapped with the model tokenizer's native chat template. We add no task-specific answer-format instruction.

Because chat templates differ across model families, we describe the prompting policy rather than hard-coding model-specific special tokens. Within each benchmark, all compared methods use the same base prompt; methods differ only in how reasoning steps are scored or selected for revision.

\subsection{Reasoning-Step Segmentation}

For natural-language reasoning tasks, reasoning steps are defined as newline-delimited non-empty text blocks. Empty spans are discarded, and if a response contains no newline boundary, the complete non-empty response is treated as a single step. Tokenizer offset mappings are used to align generated tokens with their corresponding text spans for step-level routing and logit aggregation. Character offsets are retained for each step and used to map generated tokens to their corresponding step spans.

HumanEval follows the analogous rule at the code level: each non-empty source-code line is treated as a reasoning/code step. This preserves syntactic line structure when selecting a revision point and reconstructing the preserved code prefix.

\subsection{Answer Extraction}

Task-specific answer extraction is applied consistently across Vanilla, rethink-based methods, and PRM-guided inference.

For GSM8K and MATH500, we prioritize the last well-formed \verb|\boxed{...}| expression, followed by the \verb|####| marker, ``The answer is'' patterns, and the last numerical expression in the response as a fallback. Extracted numerical strings are normalized by removing commas and canonicalizing finite integer or floating-point representations before comparison with the reference answer.

For MMLU, we extract an answer from $\{A,B,C,D\}$. For LogicBench BQA, we extract the final yes/no decision, using the last Boolean answer when necessary. The task-specific scoring implementations then compare the extracted prediction with the gold label.

\subsection{HumanEval Evaluation}

For HumanEval, model outputs are converted into executable code completions before evaluation. If a response contains a Markdown code block, its code content is extracted; duplicated benchmark prompt text is removed when present. Generation uses code-oriented stopping sequences including subsequent function or class definitions.

We evaluate the resulting completions using the official \texttt{human\_eval} functional-correctness harness and report pass@1. Evaluation uses eight worker processes and a 10-second execution timeout per completion. For rethink-based methods, the preserved code prefix and regenerated suffix are combined before execution, so correctness is always measured on the final complete program.

\subsection{IFEval Evaluation}

IFEval responses are evaluated using the official instruction-following checker under both strict and loose criteria. The evaluator records prompt-level compliance, instruction-level compliance, and per-instruction pass/fail outcomes for both variants.

The main paper reports \emph{loose prompt accuracy}. For process-supervision construction, however, we use the strict criterion to define rollout correctness and positive completion candidates, yielding a more conservative supervision signal. Loose outcomes are retained for evaluation and analysis but are not used as the primary supervision criterion.

\section{Qualitative Examples}
\label{app:qualitative}

We provide representative examples illustrating both successful and unsuccessful RoCo-Rethink revisions.

\paragraph{Successful correction (W2R).}

Table~\ref{tab:qualitative_w2r} shows a GSM8K example in which the initial response is incorrect but RoCo-Rethink eventually recovers the correct answer. The model initially predicts \$360 instead of the gold answer \$800. RoCo-Rethink selects Step 2 as the lowest-confidence revision point, corresponding to the step with confidence $C_i=-0.707$. Although the first rethink round changes the answer only from 360 to 500, a subsequent revision corrects the underlying reasoning and yields the correct answer 800; the final rethink round then preserves this answer.

\begin{table}[t]
\centering
\small
\setlength{\tabcolsep}{3pt}
\renewcommand{\arraystretch}{0.95}
\caption{\textbf{Successful RoCo-Rethink correction on GSM8K.}
The selected revision point is highlighted in bold. Routing statistics are trajectory-normalized before computing
$C_i=\widetilde M_i+\widetilde H_i-\widetilde D_i$.}
\label{tab:qualitative_w2r}
\resizebox{\linewidth}{!}{
\begin{tabular}{c p{0.58\linewidth} cccc}
\toprule
Step & Reasoning step &
$\widetilde M_i$ &
$\widetilde H_i$ &
$\widetilde D_i$ &
$C_i$ \\
\midrule
1 &
John bought twice as many red ties as blue ties, so he bought $2\times$(number of blue ties) red ties.
& 0.114 & 0.135 & 0.803 & -0.554 \\

\textbf{2} &
\textbf{The red ties cost 50\% more than blue ties, so the cost of the red ties is $1.5\times\$40=\$60$ each.}
& \textbf{0.138} & \textbf{0.155} & \textbf{1.000} & \textbf{-0.707} \\

3 &
He spent \$200 on blue ties, so he spent \$200 $-$ \$40 = \$160 on red ties.
& 0.000 & 0.000 & 0.657 & -0.657 \\

4 &
The total cost of all the ties is \$160 + \$200 = \$360.
& 0.049 & 0.138 & 0.522 & -0.335 \\

5 &
\#\#\#\# 360
& 1.000 & 1.000 & 0.000 & 2.000 \\

6 &
The answer is: 360
& 0.738 & 0.742 & 0.957 & 0.523 \\
\bottomrule
\end{tabular}}
\end{table}

\noindent
\textbf{Original suffix.}
\emph{The red ties cost 50\% more than blue ties, so the cost of the red ties is $1.5\times\$40=\$60$ each. He spent \$200 on blue ties, so he spent \$200 $-$ \$40 = \$160 on red ties. The total cost of all the ties is \$160 + \$200 = \$360.}

\noindent
\textbf{Regenerated reasoning.}
After preserving the prefix before Step 2, the model instead reasons that \$200 spent on \$40 blue ties corresponds to five blue ties, hence ten red ties. At \$60 per red tie, the red ties cost \$600, giving a total expenditure of \$800.

\noindent
\textbf{Answer trajectory.}
$360 \rightarrow 500 \rightarrow 800 \rightarrow 800$.

The example illustrates that RoCo-Rethink need not correct the trajectory in a single regeneration. The first revision alters the continuation but remains incorrect, while a subsequent low-confidence revision produces the correct reasoning path and the stopping rule terminates once the answer stabilizes.

\paragraph{Failure case (W2W).}

Table~\ref{tab:qualitative_w2w} presents a contrasting GSM8K example in which RoCo-Rethink identifies a low-confidence revision point but fails to escape the original erroneous reasoning path. The model incorrectly treats the 150\% increase as applying to the total purchase-and-repair cost, producing a final answer of \$195,000 instead of the gold profit of \$70,000. RoCo-Rethink selects Step 3 as the lowest-confidence step, with $C_i=-0.569$, but regeneration reproduces the same suffix and therefore leaves the final answer unchanged.

\begin{table}[t]
\centering
\small
\setlength{\tabcolsep}{3pt}
\renewcommand{\arraystretch}{0.95}
\caption{\textbf{RoCo-Rethink failure case on GSM8K.}
The selected revision point is highlighted in bold. Although the selector identifies the lowest-confidence step, regeneration reproduces the original erroneous continuation.}
\label{tab:qualitative_w2w}
\resizebox{\linewidth}{!}{
\begin{tabular}{c p{0.58\linewidth} cccc}
\toprule
Step & Reasoning step &
$\widetilde M_i$ &
$\widetilde H_i$ &
$\widetilde D_i$ &
$C_i$ \\
\midrule
1 &
The house was bought for \$80,000 and then had \$50,000 in repairs, so the total cost was \$130,000.
& 0.096 & 0.018 & 0.564 & -0.451 \\

2 &
The value increased by 150\%, so the new value is \$130,000 + \$195,000 = \$325,000.
& 0.170 & 0.162 & 0.742 & -0.410 \\

\textbf{3} &
\textbf{Josh sold the house for \$325,000, so his profit is \$325,000 $-$ \$130,000 = \$195,000.}
& \textbf{0.000} & \textbf{0.000} & \textbf{0.569} & \textbf{-0.569} \\

4 &
\#\#\#\# 195000
& 1.000 & 1.000 & 0.000 & 2.000 \\

5 &
The answer is: 195000
& 0.742 & 0.696 & 1.000 & 0.438 \\
\bottomrule
\end{tabular}}
\end{table}

\noindent
\textbf{Original suffix.}
\emph{Josh sold the house for \$325,000, so his profit is \$325,000 $-$ \$130,000 = \$195,000.}

\noindent
\textbf{Regenerated suffix.}
The regenerated suffix is identical to the original suffix, and the predicted answer remains 195000.

This failure case highlights an important limitation of selective rethink: identifying a low-confidence revision point does not guarantee that regeneration will leave the original reasoning basin. In this example, the generation process returns the same erroneous continuation, resulting in a W2W transition.

\end{document}

%% file: math_commands.tex
\usepackage{amsmath,amsfonts,bm}

\def\eqref#1{equation~\ref{#1}}

\def\1{\bm{1}}

\DeclareMathAlphabet{\mathsfit}{\encodingdefault}{\sfdefault}{m}{sl}
\SetMathAlphabet{\mathsfit}{bold}{\encodingdefault}{\sfdefault}{bx}{n}



%% file: tables/main_results.tex
\newcommand{\score}[2]{#1{\scriptsize $\pm$#2}}
\begin{table}[t]
\caption{\textbf{Main results of RoCo-Rethink and RoCo-PRM.}
Results are reported as mean $\pm$ SD over five random seeds.
Bold and underline indicate the best and second-best results, respectively.}
\label{tab:main_results}
\centering
\small
\setlength{\tabcolsep}{3pt}
\renewcommand{\arraystretch}{0.9}
\resizebox{\linewidth}{!}{
\begin{tabular}{llccccccc}
\toprule
Model & Method & GSM8K & MATH500 & HumanEval & MMLU & IFEval & LogicBench & Avg.\\
\midrule
\multicolumn{9}{c}{\textbf{RoCo-Rethink}}\\
\midrule
\multirow{4}{*}{OLMoE} & Vanilla & 48.7{\scriptsize $\pm$0.3} & 9.4{\scriptsize $\pm$0.5} & 29.3{\scriptsize $\pm$0.4} & 46.8{\scriptsize $\pm$0.1} & 25.7{\scriptsize $\pm$0.4} & 46.7{\scriptsize $\pm$0.3} & 34.4{\scriptsize $\pm$0.3} \\
 & Self-Consistency & 49.7{\scriptsize $\pm$0.5} & 9.8{\scriptsize $\pm$0.7} & 30.5{\scriptsize $\pm$0.7} & 47.9{\scriptsize $\pm$0.1} & 26.6{\scriptsize $\pm$0.6} & 47.2{\scriptsize $\pm$0.2} & 35.3{\scriptsize $\pm$0.5} \\
 & LECO & \underline{50.5}{\scriptsize $\pm$0.5} & \underline{11.4}{\scriptsize $\pm$0.6} & \underline{31.1}{\scriptsize $\pm$0.5} & \underline{49.0}{\scriptsize $\pm$0.1} & \underline{28.7}{\scriptsize $\pm$0.5} & \underline{55.4}{\scriptsize $\pm$0.4} & \underline{37.7}{\scriptsize $\pm$0.4} \\
 \rowcolor{rocobg}
 & RoCo-Rethink & \textbf{51.6}{\scriptsize $\pm$0.4} & \textbf{12.0}{\scriptsize $\pm$0.4} & \textbf{32.3}{\scriptsize $\pm$0.4} & \textbf{50.0}{\scriptsize $\pm$0.1} & \textbf{30.5}{\scriptsize $\pm$0.8} & \textbf{56.7}{\scriptsize $\pm$0.5} & \textbf{38.9}{\scriptsize $\pm$0.4} \\
\midrule
\multirow{4}{*}{DeepSeek} & Vanilla & 69.8{\scriptsize $\pm$0.5} & 20.6{\scriptsize $\pm$0.6} & 46.3{\scriptsize $\pm$0.5} & 49.2{\scriptsize $\pm$0.1} & 29.8{\scriptsize $\pm$0.4} & 64.6{\scriptsize $\pm$0.5} & 46.7{\scriptsize $\pm$0.4} \\
 & Self-Consistency & 70.9{\scriptsize $\pm$0.4} & 23.6{\scriptsize $\pm$0.6} & \underline{48.2}{\scriptsize $\pm$0.4} & 50.2{\scriptsize $\pm$0.1} & 30.7{\scriptsize $\pm$0.7} & 67.3{\scriptsize $\pm$0.4} & 48.5{\scriptsize $\pm$0.4} \\
 & LECO & \underline{72.7}{\scriptsize $\pm$0.6} & \underline{25.6}{\scriptsize $\pm$0.8} & \underline{48.2}{\scriptsize $\pm$0.6} & \underline{51.4}{\scriptsize $\pm$0.1} & \underline{31.4}{\scriptsize $\pm$0.5} & \underline{68.2}{\scriptsize $\pm$0.6} & \underline{49.6}{\scriptsize $\pm$0.5} \\
 \rowcolor{rocobg} & RoCo-Rethink & \textbf{74.8}{\scriptsize $\pm$0.6} & \textbf{27.0}{\scriptsize $\pm$0.5} & \textbf{49.4}{\scriptsize $\pm$0.9} & \textbf{52.5}{\scriptsize $\pm$0.1} & \textbf{32.8}{\scriptsize $\pm$0.6} & \textbf{72.4}{\scriptsize $\pm$0.3} & \textbf{51.5}{\scriptsize $\pm$0.5} \\
\midrule
\multirow{4}{*}{Qwen1.5} & Vanilla & 49.8{\scriptsize $\pm$0.6} & 16.4{\scriptsize $\pm$0.4} & 18.9{\scriptsize $\pm$0.4} & 55.3{\scriptsize $\pm$0.1} & 26.3{\scriptsize $\pm$0.3} & 63.8{\scriptsize $\pm$0.5} & 38.4{\scriptsize $\pm$0.4} \\
 & Self-Consistency & 50.7{\scriptsize $\pm$0.7} & 18.2{\scriptsize $\pm$0.3} & 21.3{\scriptsize $\pm$0.7} & 55.7{\scriptsize $\pm$0.1} & 28.1{\scriptsize $\pm$0.5} & 65.9{\scriptsize $\pm$0.6} & 40.0{\scriptsize $\pm$0.5} \\
 & LECO & \underline{52.3}{\scriptsize $\pm$0.6} & \underline{19.6}{\scriptsize $\pm$0.5} & \underline{23.2}{\scriptsize $\pm$0.5} & \underline{57.2}{\scriptsize $\pm$0.1} & \underline{28.8}{\scriptsize $\pm$0.4} & \underline{67.4}{\scriptsize $\pm$0.3} & \underline{41.4}{\scriptsize $\pm$0.4} \\
 \rowcolor{rocobg} & RoCo-Rethink & \textbf{54.6}{\scriptsize $\pm$0.4} & \textbf{20.8}{\scriptsize $\pm$0.8} & \textbf{23.8}{\scriptsize $\pm$0.4} & \textbf{58.7}{\scriptsize $\pm$0.1} & \textbf{30.7}{\scriptsize $\pm$0.6} & \textbf{68.9}{\scriptsize $\pm$0.5} & \textbf{42.9}{\scriptsize $\pm$0.5} \\
\midrule
\multicolumn{9}{c}{\textbf{RoCo-PRM}}\\
\midrule
\multirow{3}{*}{OLMoE} & MC-PRM & 50.6{\scriptsize $\pm$0.6} & 10.2{\scriptsize $\pm$0.7} & 29.9{\scriptsize $\pm$0.6} & 49.2{\scriptsize $\pm$0.1} & 26.7{\scriptsize $\pm$0.5} & 50.4{\scriptsize $\pm$0.2} & 36.2{\scriptsize $\pm$0.5} \\
 & CPMI-PRM & \underline{51.3}{\scriptsize $\pm$0.4} & \underline{10.9}{\scriptsize $\pm$0.9} & \underline{31.2}{\scriptsize $\pm$0.7} & \underline{50.3}{\scriptsize $\pm$0.1} & \underline{27.5}{\scriptsize $\pm$0.6} & \underline{56.3}{\scriptsize $\pm$0.3} & \underline{37.9}{\scriptsize $\pm$0.5} \\
 \rowcolor{rocobg} & RoCo-PRM & \textbf{53.0}{\scriptsize $\pm$0.5} & \textbf{13.4}{\scriptsize $\pm$0.7} & \textbf{32.9}{\scriptsize $\pm$0.5} & \textbf{51.3}{\scriptsize $\pm$0.1} & \textbf{29.4}{\scriptsize $\pm$0.8} & \textbf{57.8}{\scriptsize $\pm$0.3} & \textbf{39.6}{\scriptsize $\pm$0.5} \\
\midrule
\multirow{3}{*}{DeepSeek} & MC-PRM & 71.9{\scriptsize $\pm$0.8} & 24.4{\scriptsize $\pm$0.4} & 47.2{\scriptsize $\pm$0.4} & 50.5{\scriptsize $\pm$0.1} & 30.3{\scriptsize $\pm$0.8} & 69.0{\scriptsize $\pm$0.5} & 48.9{\scriptsize $\pm$0.5} \\
 & CPMI-PRM & \underline{72.2}{\scriptsize $\pm$0.4} & \underline{25.0}{\scriptsize $\pm$0.6} & \underline{47.7}{\scriptsize $\pm$0.5} & \underline{50.9}{\scriptsize $\pm$0.1} & \underline{31.5}{\scriptsize $\pm$0.4} & \underline{71.1}{\scriptsize $\pm$0.5} & \underline{49.7}{\scriptsize $\pm$0.4} \\
 \rowcolor{rocobg} & RoCo-PRM & \textbf{73.8}{\scriptsize $\pm$0.6} & \textbf{26.7}{\scriptsize $\pm$0.7} & \textbf{48.8}{\scriptsize $\pm$0.7} & \textbf{51.9}{\scriptsize $\pm$0.1} & \textbf{33.1}{\scriptsize $\pm$0.5} & \textbf{72.8}{\scriptsize $\pm$0.4} & \textbf{51.2}{\scriptsize $\pm$0.5} \\
\midrule
\multirow{3}{*}{Qwen1.5} & MC-PRM & 52.2{\scriptsize $\pm$0.3} & 19.4{\scriptsize $\pm$0.4} & 20.1{\scriptsize $\pm$0.2} & 56.5{\scriptsize $\pm$0.1} & 27.8{\scriptsize $\pm$0.7} & 66.6{\scriptsize $\pm$0.4} & 40.4{\scriptsize $\pm$0.4} \\
 & CPMI-PRM & \underline{52.9}{\scriptsize $\pm$0.8} & \underline{19.6}{\scriptsize $\pm$0.3} & \underline{22.3}{\scriptsize $\pm$0.5} & \underline{57.4}{\scriptsize $\pm$0.1} & \underline{29.4}{\scriptsize $\pm$0.5} & \underline{68.6}{\scriptsize $\pm$0.6} & \underline{41.7}{\scriptsize $\pm$0.5} \\
 \rowcolor{rocobg} & RoCo-PRM & \textbf{54.3}{\scriptsize $\pm$0.5} & \textbf{21.4}{\scriptsize $\pm$0.4} & \textbf{23.4}{\scriptsize $\pm$0.5} & \textbf{58.5}{\scriptsize $\pm$0.1} & \textbf{30.9}{\scriptsize $\pm$0.6} & \textbf{70.9}{\scriptsize $\pm$0.8} & \textbf{43.2}{\scriptsize $\pm$0.5} \\
\midrule
\multirow{4}{*}{Qwen2.5} & Vanilla & 81.2{\scriptsize $\pm$0.5} & 50.0{\scriptsize $\pm$0.5} & 49.3{\scriptsize $\pm$0.4} & 70.0{\scriptsize $\pm$0.1} & 45.2{\scriptsize $\pm$0.6} & 70.0{\scriptsize $\pm$0.5} & 61.0{\scriptsize $\pm$0.4} \\
 & MC-PRM & 82.5{\scriptsize $\pm$0.4} & 52.2{\scriptsize $\pm$0.3} & 51.2{\scriptsize $\pm$0.8} & 70.6{\scriptsize $\pm$0.1} & 46.2{\scriptsize $\pm$0.3} & 70.2{\scriptsize $\pm$0.6} & 62.2{\scriptsize $\pm$0.4} \\
 & CPMI-PRM & \underline{83.6}{\scriptsize $\pm$0.6} & \underline{53.8}{\scriptsize $\pm$0.4} & \underline{52.4}{\scriptsize $\pm$0.4} & \textbf{70.7}{\scriptsize $\pm$0.1} & \underline{46.8}{\scriptsize $\pm$0.4} & \underline{70.3}{\scriptsize $\pm$0.4} & \underline{62.9}{\scriptsize $\pm$0.4} \\
 \rowcolor{rocobg} & RoCo-PRM & \textbf{85.0}{\scriptsize $\pm$0.9} & \textbf{55.4}{\scriptsize $\pm$0.5} & \textbf{54.3}{\scriptsize $\pm$0.7} & \textbf{70.7}{\scriptsize $\pm$0.1} & \textbf{48.1}{\scriptsize $\pm$0.5} & \textbf{72.1}{\scriptsize $\pm$0.6} & \textbf{64.3}{\scriptsize $\pm$0.6} \\
\midrule
\multirow{4}{*}{Llama} & Vanilla & 76.3{\scriptsize $\pm$0.3} & 29.8{\scriptsize $\pm$0.4} & 47.6{\scriptsize $\pm$0.2} & 60.5{\scriptsize $\pm$0.1} & 47.5{\scriptsize $\pm$0.3} & 64.7{\scriptsize $\pm$0.2} & 54.4{\scriptsize $\pm$0.3} \\
 & MC-PRM & 79.2{\scriptsize $\pm$0.4} & 31.0{\scriptsize $\pm$0.7} & 49.4{\scriptsize $\pm$0.5} & \underline{62.1}{\scriptsize $\pm$0.1} & 48.8{\scriptsize $\pm$0.5} & 68.5{\scriptsize $\pm$0.4} & 56.5{\scriptsize $\pm$0.4} \\
 & CPMI-PRM & \underline{79.8}{\scriptsize $\pm$0.6} & \underline{32.6}{\scriptsize $\pm$0.5} & \underline{50.0}{\scriptsize $\pm$0.5} & \underline{62.1}{\scriptsize $\pm$0.1} & \underline{49.4}{\scriptsize $\pm$0.4} & \underline{69.7}{\scriptsize $\pm$0.8} & \underline{57.3}{\scriptsize $\pm$0.5} \\
 \rowcolor{rocobg} & RoCo-PRM & \textbf{81.4}{\scriptsize $\pm$0.7} & \textbf{34.2}{\scriptsize $\pm$0.6} & \textbf{51.8}{\scriptsize $\pm$0.8} & \textbf{63.4}{\scriptsize $\pm$0.1} & \textbf{51.3}{\scriptsize $\pm$0.6} & \textbf{71.5}{\scriptsize $\pm$0.5} & \textbf{58.9}{\scriptsize $\pm$0.6} \\
\bottomrule
\end{tabular}}
\end{table}